\documentclass[journal,10pt,twocolumn]{IEEETran}

\usepackage{threeparttable}
\usepackage{cite}
\usepackage{amsmath,amssymb,amsfonts}
\usepackage{algorithmic}
\usepackage{graphicx}
\usepackage{textcomp}
\usepackage{indentfirst}
\usepackage{csquotes}
\usepackage{multirow}
\usepackage{tabularx}
\usepackage{float}
\usepackage{color}
\usepackage{comment}
\usepackage{caption2}
\usepackage{algorithm2e}

\begin{document}

\title{Crossbar-Aware Neural Network Pruning}

\author{Ling Liang, Lei Deng, \IEEEmembership{Member}, \IEEEmembership{IEEE}, Yueling Zeng, Xing Hu, Yu Ji, Xin Ma, \\Guoqi Li, \IEEEmembership{Member}, \IEEEmembership{IEEE},  and Yuan Xie, \IEEEmembership{Fellow}, \IEEEmembership{IEEE}\\

 \thanks{Ling Liang and Lei Deng contributed equally to this work, corresponding authors: Yuan Xie and Guoqi Li. The work was partially supported by National Science Foundation (Grant No. 1719160, 1725447 and 1730309) and National Natural Science Foundation of China (Grant No. 61603209 and 61876215). Ling Liang, Lei Deng, Yueling Zeng, Xing Hu, Xin Ma and Yuan Xie are with the Department of Electrical and Computer Engineering, University of California, Santa Barbara, CA 93106, USA (email: lingliang@ucsb.edu, leideng@ucsb.edu, yuelingzeng@ucsb.edu, xinghu@ucsb.edu, xinma@ucsb.edu and yuanxie@ucsb.edu). Guoqi  Li is with the Department  of Precision Instrument, Center for Brain Inspired Computing Research, Tsinghua University, Beijing 100084, China  (email: liguoqi@mail.tsinghua.edu.cn). Yu Ji is with the Department of Computer Science and Technology, Tsinghua University, Beijing 100084, China  (email: jiy15@mails.tsinghua.edu.cn). }} % stops a space

\maketitle

\begin{abstract}

Crossbar architecture has been widely adopted in neural network accelerators due to the efficient implementations on vector-matrix multiplication (VMM) operations. However, in the case of convolutional neural networks (CNNs), the efficiency is compromised dramatically because of the large amounts of data reuse. Although some mapping methods have been designed to achieve a balance between the execution throughput and resource overhead, the  resource consumption cost  is still huge while maintaining the throughput. 

Network pruning is a promising and widely studied method to shrink the model size. Whereas, prior work for CNNs compression rarely considered the crossbar architecture and the corresponding mapping method, and cannot be directly utilized by crossbar-based neural network accelerators. This paper proposes a crossbar-aware pruning framework based on  a formulated  $L_0$-norm constrained optimization problem. Specifically, we design an $L_0$-norm constrained gradient descent (LGD) with  relaxant probabilistic projection (RPP) to solve this problem. Two types of sparsity are successfully achieved: i) intuitive crossbar-grain sparsity and  ii) column-grain sparsity with output recombination, based on which we further propose an input feature maps (FMs)  reorder method to improve the model accuracy. We evaluate our crossbar-aware pruning framework on the median-scale CIFAR10 dataset and the large-scale ImageNet dataset with VGG and ResNet models. Our method is able to reduce the crossbar overhead  by 44\%-72\% with insignificant accuracy degradation. This work significant reduce the resource overhead and the related energy cost, and provides a new co-design solution for mapping CNNs onto various crossbar devices with much better efficiency. 

\end{abstract}

{ \it Keywords:}  Crossbar Architecture, Convolutional Neural Networks,  Neural Network Pruning, Constrained Optimization Problem

\section{Introduction}\label{sec:introduction}
It is well known that crossbar with analog-domain computing naturally boosts the performance of vector-matrix multiplication (VMM), which is the major operation in existing neural networks (NNs) \cite{chi2016prime,li2013memristor,hu2012hardware}. For this reason, nowadays crossbar architecture based on conventional memory devices (e.g. SRAM \cite{zhang2017memory, merolla2014million, 7409624, davies2018loihi} and Flash \cite{guo2017fast}) or emerging memory devices (e.g. RRAM \cite{garbin2014variability, deng2015complex, long2016reram, chi2016prime, shafiee2016isaac, tang2017binary, song2017pipelayer}, PCRAM \cite{ambrogio2018equivalent, burr2015experimental, bichler2012visual}, MRAM \cite{fan2017energy}, etc.) are being widely used in neural network (NN) accelerators. A functional crossbar is a self-contained small NN with a weighted connection crossbar and its peripheral units. Many functional crossbars are wired by a 2D scalable routing infrastructure, and this forms a massively parallel so-called \textit{many-crossbar architecture}. Based on a variety of inter-crossbar communication topologies, such as tree\cite{merolla2014multicast}, 2D triangular toroidal mesh \cite{painkras2013spinnaker} or 2D XY mesh \cite{merolla2014million, akopyan2015truenorth}, many-crossbar architecture has  demonstrated high performance on various NN benchmarks compared to traditional platforms.
% * <liguoqi@mail.tsinghua.edu.cn> 2018-07-18T09:02:30.328Z:
%
% ^.

Even though the many-crossbar architecture usually performs well in multi-layered perceptron (MLP) with dense VMM operations, the efficiency is compromised on convolutional neural networks (CNNs) due to the large amounts of data reuse. To map CNNs onto the many-crossbar  architecture, fully-unfolding the reused data is a straightforward solution \cite{esser2016convolutional, ji2018bridge}. Specifically, it unfolds the reused neuron activations and synaptic weights, and assigns independent resources for all these unfolded cells. It can achieve extremely throughput compared to the fully-folded mapping that reuses all the neurons and weights \cite{chi2016prime} cycle by cycle, however, this scheme consumes significantly huge crossbar resources. For example, more than 3$\times$10$^4$ crossbars are needed to achieve comparable accuracy on medium-scale CIFAR-10 dataset \cite{krizhevsky2009learning} even if a network pruning approach has been leveraged \cite{esser2016convolutional}; more than hundreds of thousands of crossbars \cite{ji2018bridge} are often occupied for larger models \cite{krizhevsky2012imagenet, simonyan2014very, he2016deep} on ImageNet dataset \cite{deng2009imagenet}. Although a compromised solution, termed as semi-folded mapping \cite{deng2018semi}, has emerged to achieve a balance between the execution throughput and resource overhead, the resource and the resulting energy cost are still very high. This becomes a fatal blow for the compact and energy-limited embedded devices.

Recently, many works on NN pruning seem promising to shrink large models to reduce the resource and energy consumption. In detail, various sparsity structures from the element grain \cite{han2015learning, han2015deep}, element group grain \cite{yu2017scalpel}, vector grain \cite{wen2016learning, wen2017learning}, and even channel grain \cite{he2017channel,li2016pruning,molchanov2016pruning,luo2017thinet}, have been proposed. On one side, although the sparse network obtained from channel-wise pruning can be reorganized as a compact dense model for any architecture, the accuracy loss is usually high due to the aggressive sparsity. On the other side, almost all the smaller-grain pruning works consider only the original logic computational graph or the execution graph on general purpose platforms (e.g. CPU/GPU), instead of the execution graph after mapping onto the crossbar architecture. Although it is possible, also widely used, to save energy via adding compute gate benefit from the vast zeros in a sparse network, the crossbar still cannot be removed due to the residual irregular non-zeros. Therefore, we have to leave its costly peripheral circuits. It is difficult to efficiently leverage these fine-grained sparse models in practical crossbar devices. In this sense, reducing the number of crossbars as much as possible is the most effective way to save the resource and energy cost. With this guidance, previous work attempted to obtain crossbar-oriented pruning using iterative clustering method \cite{ankit2017trannsformer}, but the fully-connected (FC) layer was the focus. The effective pruning of Conv layer on crossbars is still a remaining challenge, since the Conv mapping is more complex than the intuitive FC mapping due to the data reuse and the pruning difficulty is increased due to the less redundancy.

Motivated by the above analysis, we try to answer one important yet untouched question, i.e. how many crossbars can we reduce when mapping CNNs onto the crossbar architecture? To this end, a crossbar-aware pruning framework is proposed to minimize the resource overhead on crossbar architecture for CNNs. Firstly, two crossbar-friendly sparsity grains, crossbar grain and column grain, are designed based on the semi-folded mapping method. The former sparsity is straightforward and easy to map, and the latter one can be converted to the crossbar-grain sparsity by recombining non-zero columns along the output dimension. Secondly, we formulate the crossbar-aware pruning as an $L_0$-norm constrained optimization problem, and propose an $L_0$-norm constrained gradient descent (LGD) with relaxant probabilistic projection (RPP) method to solve it. Thirdly, a reorder of input FMs is proposed to enhance the model accuracy. We conduct a comprehensive analysis on the resource-accuracy trade-off on various benchmarks, including medium-scale CIFAR10 and large-scale ImageNet datasets with VGG and ResNet models. Overall, our crossbar-aware pruning framework is efficient for crossbar architecture, which is able to reduce 44\%-72\% crossbar overhead with acceptable accuracy degradation. This paper provides a new co-design solution for mapping CNNs onto various crossbar devices with significantly higher efficiency. 

The contributions of this work are briefly summarized as follows:
\begin{itemize}
\item{We propose an effective pruning method for CNN implementation on crossbars, which can significantly reduce the resource overhead.}

\item{The LGD solver is able to control the resulting sparsity accurately, and the probabilistic projection helps improve the algorithm convergence.}

\item{The effectiveness of our method was evaluated on various datasets, including the large-scale ImageNet.}

\end{itemize}

The rest of this paper is organized as follows: Section II introduces the crossbar architecture and the mapping strategy; Section III explains the details of our crossbar-aware neural network pruning framework; the technique analysis and performance evaluations are reported in Section IV ; finally, Section V concludes and discusses this paper. 
\section{Crossbar architecture and CNN Mapping}\label{sec:Backgrounds}
\begin{figure}[!htbp]
\centering
\includegraphics[width=0.48\textwidth]{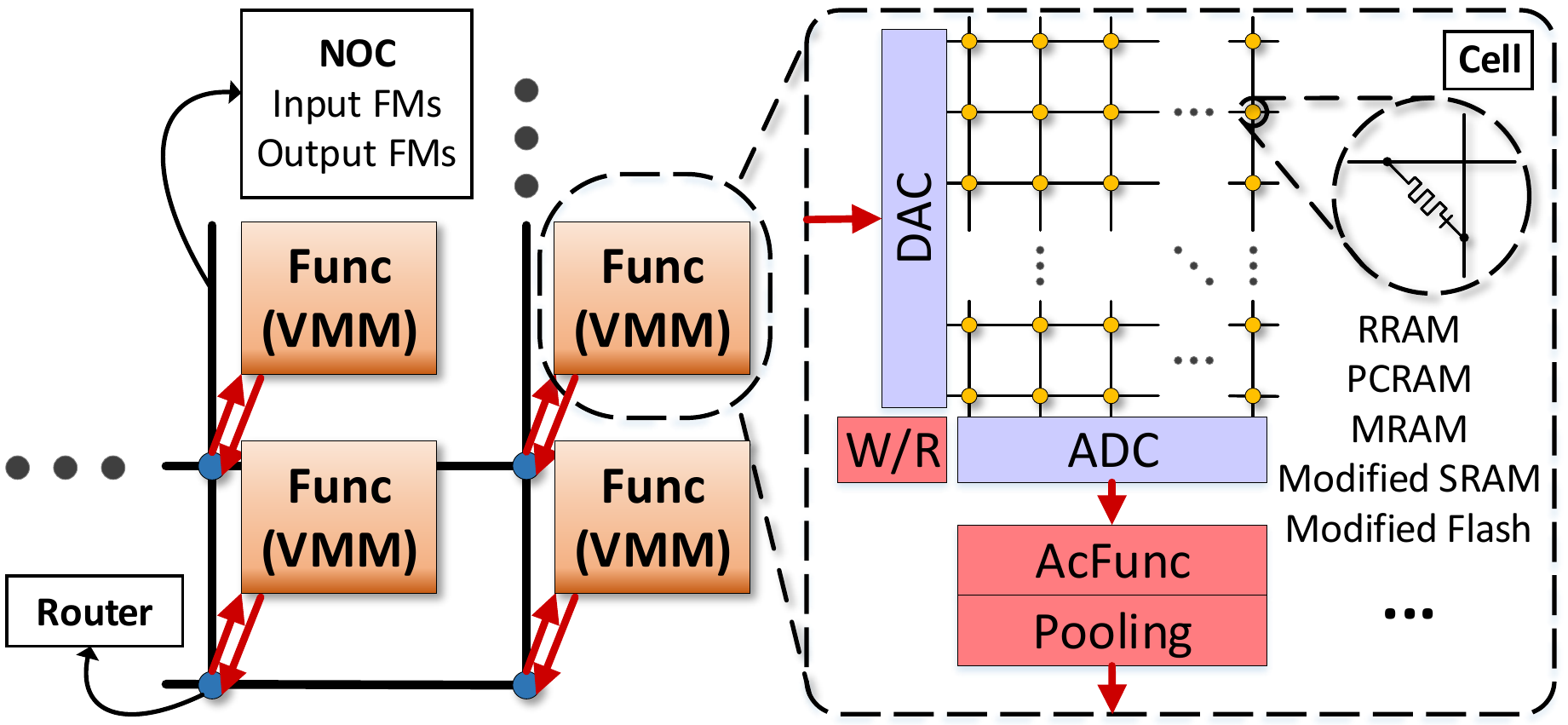}
\caption{Overview of crossbar-based architecture.}
\label{crossbar}
\vspace{-10pt}
\end{figure}

Normally, the crossbar-based architecture is a hierarchical and scalable system. In such architecture, a basic building block, called functional crossbar (Func), consists of a crossbar and its peripheral circuits (e.g. read/write driver, ADC/DAC, activation function and pooling circuits, timing scheduler, etc.) and can efficiently perform the vector-matrix multiplication (VMM). As shown in Figure \ref{crossbar}, the Func units are connected by a routing structure to realize communication with each other.

\subsection{Many-Crossbar Architecture}
The key component in each Func is the crossbar array. The weight parameters program the conductance of the memory array. After this mapping initialization, the input neuronal activations are injected into the crossbar word lines, and an in-place vector-matrix multiplication (VMM) can be performed by applying voltages to each row and reading currents from each column. Specifically, crossbar computing is in analog domain, thus many peripheral circuits are required for the complete functionality. For example, write and read driving circuits are used to program the memory cell and read the output state, respectively. Digital-to-analog converter (DAC) and analog-to-digital converter (ADC) switch signal format between the crossbar analog domain and the peripheral digital domain. In addition, additional digital circuits are required for activation function, pooling operation, and the timing schedule. Note that the memory cell can have many variants, including conventional memory devices with specialized modification (e.g. SRAM \cite{zhang2017memory, merolla2014million, 7409624, davies2018loihi} and Flash \cite{guo2017fast}) or emerging non-volatile memory (NVM) devices with different material hierarchy and resistive mechanism (e.g. RRAM \cite{garbin2014variability, deng2015complex, long2016reram, chi2016prime, shafiee2016isaac, tang2017binary, song2017pipelayer}, PCRAM \cite{ambrogio2018equivalent, burr2015experimental, bichler2012visual}, and MRAM \cite{fan2017energy}). Finally, each Func is equipped with a router for inter-crossbar communication which guarantees the scalability to a large-scale system.

%variable definition
\begin{table*}[!htbp]
\caption{Variable definition.}
\vspace{5pt}
\label{tab:Variables}
\centering
\renewcommand\arraystretch{1.3}
\resizebox{0.95\textwidth}{!}{
\begin{tabular}{|| c|c || c|c ||} 
\hline
Variable & Definition & Variable & Definition\\
\hline
$P,~p$ & number, index of input FM    & $I,~i$ & number, index of input FM group \\ 
$Q,~q$ & number, index of output FM   & $J,~j$ & number, index of output FM group \\ 
\hline
$g^i_{in}$   & set of input FMs{'} indexes in $i$-th input FM group   & $K_{in}$  & number of FMs in each input FM group\\
$g^j_{out}$  & set of output FMs{'} indexes in $j$-th output FM group & $K_{out}$ & number of FMs in each output FM group\\
\hline
$\pmb{W}^{p, q}$ & weight kernel connecting $p$-input and $q$-output FM & $\beta^{i,j}_{L_0}$   & pruning mask value involving $i$-th input and $j$-th output FM group\\
$\pmb{W}^q$     & weight tensor of $q$-th output FM                    & $\pmb{\beta}^j$, $\pmb{\beta}^j_{L_0}$ & pruning coefficient, mask vector involving $j$-th output FM group\\
$\pmb{X}^{p}$   & $p$-th input FM  & $r$, $r_0$   & $L_0$ norm of $\pmb{\beta}^j_{L_0}$, relaxation factor in RPP\\
\hline
$\pmb{Y}_S^{p,q}$ & initial convolution result of $\pmb{X}^p \circledast \pmb{W}^{p,q}$ &
$\pmb{Y}_{C1}^{i,j}$ & set of $\pmb{Y}_E^{i,q}$ involving $j$-th output FM group\\
$\pmb{Y}_E^{i,q}$ & partial summation of $\pmb{Y}_S^{p,q}$ involving $i$-th input FM group and $q$-th output FM&
$\pmb{Y}_{C2}^{j}$   & set of $\pmb{Y}_O^q$ involving $j$-th output FM group\\
$\pmb{Y}_O^{q}$ & Complete summation of $\pmb{Y}_S^{p,q}$ involving $q$-th output FM & $N$ & number of image samples\\
\hline
$\pmb{FM}_{L_X}$ & Tensor representation of input FMs for linear regression  & $\pmb{FM}_{Y_{C1}^j}$ & concatenation of all tensors in $\pmb{Y}_{C1}^{i,j},~i=0,~1,...,~I-1$ \\
$\pmb{FM}_{L_Y}$ & Tensor representation of output FMs for linear regression & $\pmb{FM}_{Y_O^j}$ & concatenation of all tensors in $\pmb{Y}_O^j$\\
\hline
\large $\circledast$ & convolution operation & $S_{out}$ & the size of output FM\\
\hline
\end{tabular}}
\end{table*}

\begin{comment}
importance of $p$-th input FM (summation of $\pmb{Y}_S^{p,q}$ involving $p$-th input FM) & $N$ & number of image samples
\end{comment}

\begin{figure*}[!htbp]
\centering
\includegraphics[width=0.95\textwidth]{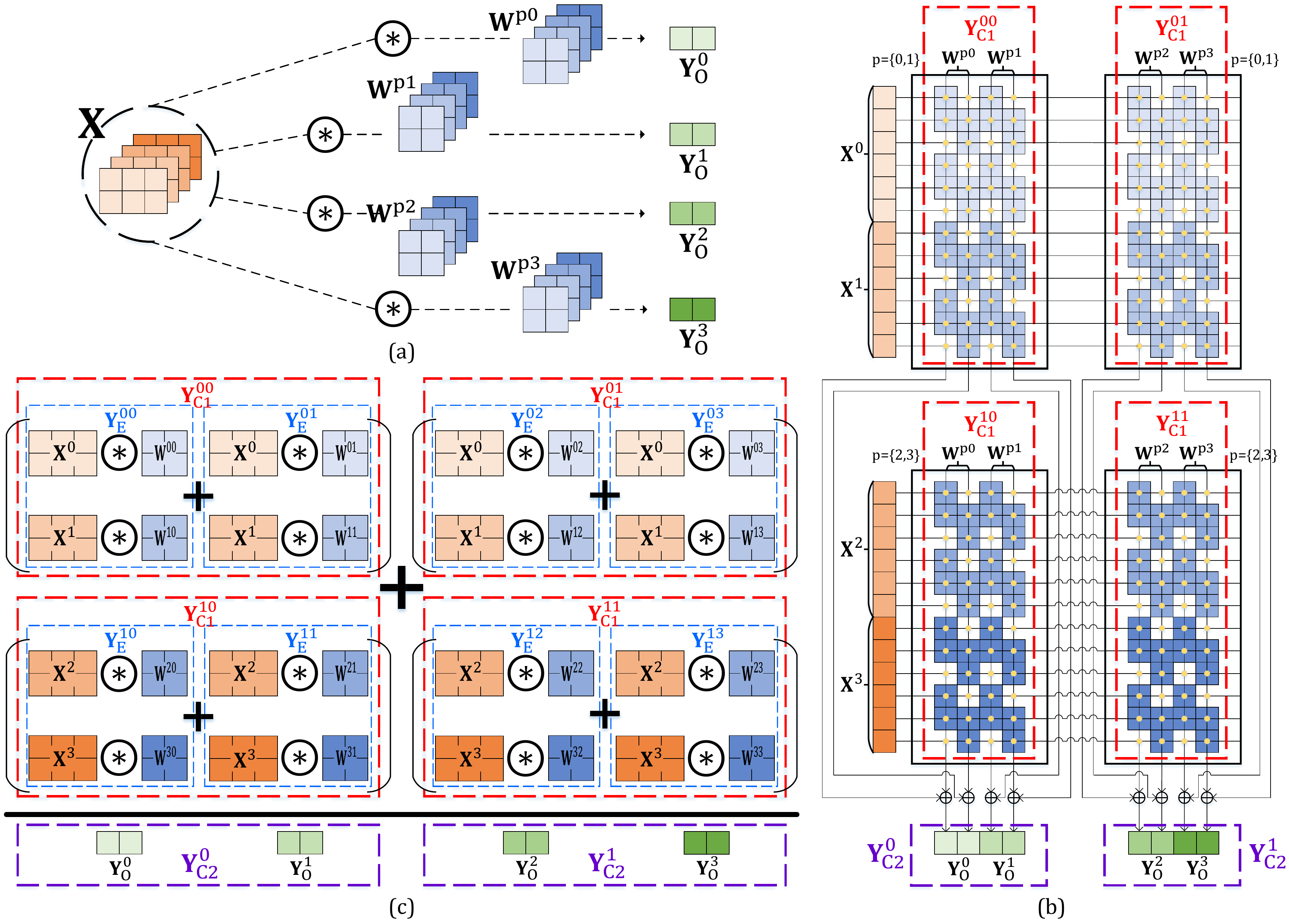}
\caption{Illustration of mapping convolutional neural networks onto crossbars: (a) a single convolutional layer; (b) semi-folded mapping on crossbar-based architecture; (c) matrix representation of each crossbar.}
\label{CNN}
\end{figure*}

\subsection{Semi-folded Mapping of Convolution Layer}
Because the crossbar architecture can improve the VMM performance, the FC layer can gain benefits naturally. However, some operations in neural network such as convolution cannot be directly deployed due to the large data reuse. Conventionally, there exist two schemes for convolution mapping on the crossbar architecture: fully-unfolded \cite{esser2016convolutional, ji2018bridge} and fully-folded \cite{chi2016prime}. The former one that is widely used in neuromorphic field unfolds all the memory and computation, and then transforms them to VMM for crossbar execution. On the other hand, the latter one only assigns the physical resources for one sliding window, then it reuses these resources cycle by cycle until all the sliding windows are finished. Overall, the fully-unfolded mapping consumes large amount of resources to achieve high throughput, while the fully-folded mapping consumes plenty of clock cycles to achieve minimum resources. In general, the resulting speed and overhead are
greatly imbalanced. To address these issues, a semi-folded mapping \cite{deng2018semi} is proposed recently, which simultaneously folds the physical resources along the row dimension of feature maps (FMs) for resource saving and unfolds them along the column dimension for maintaining parallelism. Therefore, it can balance performance and overhead to a great extent. However, as mentioned earlier, it still consumes lots of resources. In this paper, we implement our pruning methodology based on the semi-folded mapping scheme to further reduce the overhead.

Since the mapping of FC layer is much easier, we focus on the illustration of Conv mapping in this section. But note that it is quite easy to extend our method to FC layer. The variable definitions are listed in Table \ref{tab:Variables}. Figure \ref{CNN}(a) presents an example of a Conv layer, where both the number of input and output FMs are four, i.e. $P=Q=4$. The size of each input and output FM is $2 \times 3$ and $1 \times 2$, respectively. Here we just take this as an example. In fact the FM size can be much larger in semi-folded mapping. Specifically, the FM height has no limitation since the crossbar is reused along the FM row direction, and FM with larger width can be split onto independent crossbars for execution parallelism. We use $\pmb{X}$ (orange rectangles) and $\pmb{Y}$ (green rectangles) to denote the input and output FM tensors (here $\pmb{X} \in R^{2 \times 3 \times 4}$ and $\pmb{Y} \in R^{1 \times 2 \times4}$), respectively. In Conv layer, weight (blue rectangles) is a four dimensional tensor, in this example we have $\pmb{W} \in R^{2 \times 2 \times 4 \times 4}$. Each output FM $\pmb{Y}_O^q$ is the convolution result between the input FM tensor $\pmb{X}$ and the weight filters $\pmb{W}^q$.

Figure \ref{CNN}(b) illustrates the semi-folded mapping of this example. We assume that the size of each crossbar is $12 \times 4$ (12 rows and 4 columns) for simplicity. According to the mapping results, we can divide the input and output FMs into two groups $I=J=2$, such that each input-output group pair occupies one crossbar. We use $i$ and $j$ to indicate the group index (here $i \in \{0,1\}$ and $j \in \{0,1\}$). Each crossbar just generates the partial sum of the intra-group output FMs, and the additional accumulation of the results from the crossbars in one column can obtain the complete activations of output FMs. In this example, totally four crossbars are required to implement the Conv layer. 

Figure \ref{CNN}(c) shows the matrix representation of the crossbar mapping, where it is seen that  each crossbar realizes eight convolution sliding windows, and each sliding window covers a single input FM with one of its weight kernel. And each crossbar  generates a set of partial sums corresponding to its output FM group.  Then we present all the corresponding calculations involved in  Fig.2 as follows
\begin{equation} \label{initial convolution}
	\pmb{Y}_S^{p,q} = \pmb{X}^p \circledast \pmb{W}^{p,q}
\end{equation}
\begin{equation} \label{partial_complete_sum}
	\pmb{Y}_E^{i,q} = \sum\nolimits_{p \in g_{in}^i} \pmb{Y}_S^{p,q},~\pmb{Y}_O^q = \sum\nolimits_{i} \pmb{Y}_E^{i,q} 
\end{equation}
\begin{equation} \label{partial_complete_sum_set}
	\pmb{Y}_{C1}^{i,j} = \left\{ \pmb{Y}_E^{i,q} ~|~ q \in g_{out}^j\right\},~\pmb{Y}_{C2}^j = \left\{ \pmb{Y}_O^q ~|~ q \in g_{out}^j\right\}
\end{equation}
where $\pmb{Y}_E^{i,q}$ accumulates the initial convolution result $\pmb{Y}_S^{p,q}$ in the same input FM group, and we collect the indexes of input FMs in $i$-th group to as an index set $g_{in}^i$. Furthermore, $\pmb{Y}_O^{q}$ accumulates $\pmb{Y}_E^{i,q}$ across all input FM groups to generate a complete sum corresponding to the $q$-th output FM. In short, the complete sum of each output FM is the summation of all partial sums from its corresponding crossbars. Note that the data size of $\pmb{Y}_S^{p,q}$ and $\pmb{Y}_E^{i,q}$ are identical, which are same with the size of output FM, $\pmb{Y}_S^{p,q} \in R^{S_{out}}$, $\pmb{Y}_E^{i,q} \in R^{S_{out}}$. Here $S_{out}$ usually equals to FM height multiplying FM width. Based on the grouping of input FMs on crossbars, the initial convolution result $\pmb{Y}_S \in R^{S_{out} \times P \times Q}$ shrinks to partial sums $\pmb{Y}_E \in R^{S_{out} \times I \times Q}$. Equation (\ref{partial_complete_sum_set}) organizes the tensors of partial sum and complete sum in the $j$-th output FM group as tensor sets, which represent all the partial sums from the ($i$, $j$)-th crossbar as $\pmb{Y}_{C1}^{i,j}$ and the complete sums involving the $j$-th output FM group as $\pmb{Y}_{C2}^j$. The index set of output FMs in $j$-th output FM group is denoted as $g_{out}^j$.
\section{Crossbar-aware Pruning Framework}\label{sec:Partition}
In this section, we will present a comprehensive analysis of how to explore the crossbar-aware sparsity with two grain levels: crossbar-grain and column-grain. Then we formulate our pruning problem as an $L_0$-norm constrained optimization problem and propose an $L_0$-norm constrained gradient descent (LGD) to solve it. Finally, we introduce our input FM reorder method to improve the model accuracy of sparse networks.

\subsection{Crossbar-grain and Column-grain Sparsity}\label{sec:sparsity grain}
In Conv layer, the complete sum on one output FM is the summation of its partial sums produced by the convolution between all input FMs and its corresponding weight filter. The sparsity analysis is to identify that which partial sums contribute less and then prune them. After that, the remained partial sums are used to approximate the original output FMs. In this work, we design two pruning grains: crossbar-grain sparsity and column-grain sparsity which provide usable sparsity for the crossbar architecture.  

\begin{figure*}[!htbp]
\centering
\includegraphics[width=0.95\textwidth]{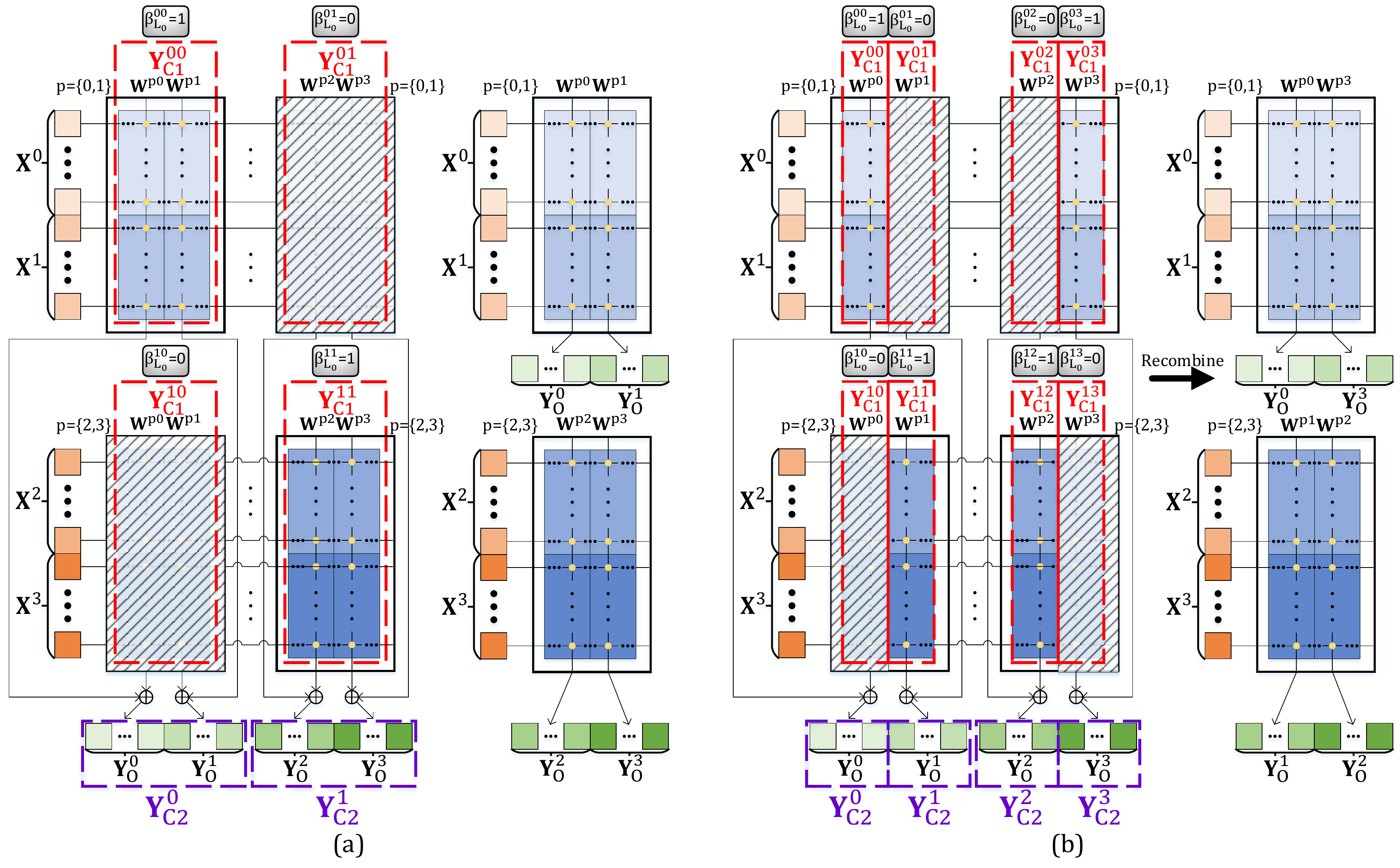}
\caption{Crossbar-aware pruning: (a) crossbar-wise pruning without recombination; (b) column-wise pruning with recombination.}
\label{pruning}
\end{figure*}

In the crossbar-grain sparsity, our goal is to eliminate crossbars whose partial sum $\pmb{Y}_{C1}^{i,j}$ contributes less to its output FM group $\pmb{Y}_{C2}^j$. We design a binary pruning mask $\beta_{L_0}^{i,j}$ to indicate whether the crossbar will be pruned ($\beta_{L_0}^{i,j}=0$) or not. Figure \ref{pruning}(a) shows an example of crossbar-grain pruning. In this case, the right top crossbar and the left bottom crossbar are pruned (marked as shadow). In detail, we initially have
\begin{equation} \label{initial convolution}
	\pmb{Y}_O^q=\pmb{Y}_E^{0,q}+\pmb{Y}_E^{1,q}, q=0,~1,~2,~3
\end{equation}
After pruning, each output FM can be calculated from the remained input FM group, i.e.
\begin{equation}
\begin{cases}
\pmb{Y}_O^0=\beta_{L_0}^{0,0}\pmb{Y}_E^{0,0},~~\pmb{Y}_O^1=\beta_{L_0}^{0,0}\pmb{Y}_E^{0,1}\\    \pmb{Y}_O^2=\beta_{L_0}^{1,1}\pmb{Y}_E^{1,2},~~\pmb{Y}_O^3=\beta_{L_0}^{1,1}\pmb{Y}_E^{1,3}
\end{cases}
\end{equation}
Then we can formulate the approximated function after pruning as follows
\begin{equation} \label{crossbar prune}
	\pmb{Y}_O^q = \sum\nolimits_{i}\beta_{L_0}^{i,j}\pmb{Y}_E^{i,q},~q\in g_{out}^j.
\end{equation}
In contrast to original dense accumulation, the accumulation of the partial sum from each crossbar is controlled by a binary pruning mask $\beta_{L_0}^{i,j}\in \{0,~1\}$. Each $\beta_{L_0}^{i,j}$ is shared by all partial sums $\pmb{Y}_E^{i,q}$ in the same $j$-th output FM group, which is usually in the same crossbar.

Although the crossbar-grain sparsity can intuitively eliminate redundant crossbars, the output FM group after pruning may stray from original output FM group a lot, if the dependency among elements $\pmb{Y}_E^{i,q}$ in $\pmb{Y}_{C1}^{i,j}$ is poor. Considering the resource-accuracy tradeoff issue, a straightforward method is to shrink the size of $K_{out}$ which is determined by the crossbar size in crossbar-grain sparsity. By reducing the number of output FMs in each output FM group, the dependency requirement among output FMs in the same output FM group can be mitigated. Thus, we further propose column-grain pruning to decrease the sparsifying grain for error reduction, which is shown in Figure \ref{pruning}(b). Now, each output FM group only contains one output FM, i.e. $J=Q=4$ and $K_{out}=1$.  After pruning, each crossbar usually has non-zero columns, not fully pruned like that in crossbar-grain pruning. In the example, the first input FM group $\pmb{X}^{\{0,1\}}$ contributes to the first output FM group $\pmb{Y}_{C2}^0$ and the fourth output FM group $\pmb{Y}_{C2}^3$. So the first half columns in the left top crossbar can be recombined with the last half columns in the right top crossbar to form a new crossbar. The input of this new crossbar is still the first input FM group $\pmb{X}^{\{0,1\}}$, whereas, after recombination the crossbar output is the first output FM group $\pmb{Y}_{C2}^0$ and the fourth output FM group $\pmb{Y}_{C2}^3$. The crossbars involving the second input FM group can be shrunk by using the same pruning and recombination method. Furthermore, if each output FM has multiple non-pruned crossbars to receive its input FMs, the inter-crossbar accumulation of the partial sums is still required (Figure \ref{pruning} omits this case for clarity). Usually, the column-grain pruning can achieve the similar sparsity with the crossbar-grain pruning, but with significantly higher accuracy since the pruning is more elaborate.

\subsection{$L_0$-norm Constrained Optimization Problem}
In previous section we have analyzed the sparsity for crossbar architecture, in which we expect the summation of remained partial sums can approximate to the original $\pmb{Y}_{C2}^{j}$ as closely as possible. For the convenience of expression, here we concatenate all the complete sums in $\pmb{Y}_{C2}^j$ into a tensor and reshape it to a vector, then denote it as $\pmb{Y} \in R^{S_{out} K_{out}}$. $\pmb{X} \in R^{S_{out} K_{out} \times I}$ is also obtained after similar concatenation and reshaping of all the partial sum sets $\pmb{Y}_{C1}^{i,j}$ involving $j$-th output FM group. In this section we simplify the number of image samples to 1 for clarity, i.e. $N=1$. The objective function of the pruning optimization problem of the $j$-th output FM group can be described as
\begin{equation} \label{L0 norm}
	\underset{\pmb{\beta}_{L_0}^j}{\text{min}} ~\|\pmb{Y} - \pmb{X}\pmb{\beta}_{L_0}^j\|_2^2 ~~~ s.t. ~\|\pmb{\beta}_{L_0}^j\|_0 = r,~\beta_{L_0}^{ij} \in\{0,~1\}
\end{equation}
where $r$ is the $L_0$ norm (number of non-zero elements, i.e. $\| \cdot \|_0$) of the binary pruning mask $\pmb{\beta}_{L_0}^j$, which determines the sparsity. The loss is the square of Euclidean distance between complete sums $\pmb{Y}$ and the sparsified partial sums $\pmb{X}\pmb{\beta}_{L_0}^j$. After transforming the pruning issue into an $L_0$-norm constrained optimization problem, the object now is to find a binary pruning mask $\pmb{\beta}_{L_0}^j$ that satisfies the $L_0$-norm constraint and minimizes the distance. However, the $L_0$-norm constrained problem is an NP problem. In this work we propose $L_0$-norm constrained gradient descent (LGD) to solve it.

\makeatletter
\renewcommand{\@algocf@capt@plain}{above}% formerly {bottom}
\makeatother

%alg1 RPP
\begin{algorithm}
\caption{Relaxant Probabilistic Projection (RPP)}
\label{RPP}
\KwData{$\pmb{\beta}^j$, $L_0$ norm $r$, relaxation factor $r_0$}
\KwResult{$\pmb{\beta}^j_{L_0}$ with $r$ nonzero elements}
\BlankLine
\BlankLine
$\pmb{\beta}_{abs}^j \leftarrow abs(\pmb{\beta}^j)$;\\
\BlankLine
// select $r+r_0$ candidates with highest importance\\
$Canditates=sort\_descend(\pmb{\beta}_{abs}^j)[0:r+r_0-1]$;\\
\For{all $i$}{
	$\beta_{abs}^{ij} \leftarrow 0, \ if \ \beta_{abs}^{ij} \notin Canditates$;
}
// select $r$ elements from the candidates through \\
// iterative probabilistic sampling\\
$\pmb{\beta}^j_{L_0}=\pmb{0}$;\\
\While{$\|\pmb{\beta}_{L_0}^j\|_0 < r$}{
	// update sampling probability by normalizing $\pmb{\beta}_{abs}^j$\\
    \For{all $i$}{
    	$\beta_{prob}^{ij}=\beta_{abs}^{ij}/\sum_{i^{'}}{\beta_{abs}^{i^{'}j}}$;\\
    }
	// probabilistically sample at one iteration, and update \\
	// $\pmb{\beta}_{abs}^j$ by removing the selected elements\\
	$Samples=\{\beta^{ij}\ |\ \beta_{prob}^{ij} > rand[0,1]\}$;\\
	\For{all $i$}{
        $\beta_{L_0}^{ij} \leftarrow \beta^{ij}$ then $\beta_{abs}^{ij} \leftarrow 0, \ if \ \beta^{ij} \in Samples$;\\
        break, \ if \ $\|\pmb{\beta}_{L_0}^j\|_0 = r$;
    	}
    }
\BlankLine
\end{algorithm}

Before introducing the whole flow of how the LGD works, we discuss a relaxant probabilistic projection (RPP) method first. The RPP can transfer a pruning coefficient vector $\pmb{\beta}^j\in R^I$ to the corresponding pruning mask $\pmb{\beta}^j_{L_0}$ whose $L_0$ norm is $r$ (not binary at this stage). Naturally, the largest $r$ elements in $|\pmb{\beta}^{j}|$ indicates the most contributions. However, this intuitive and deterministic selection is so dictatorial that completely cuts off the opportunity of the ones that are out of the largest $r$ elements but still large enough. Therefore, we introduce a relaxation factor $r_0$ and a probabilistic projection method to determine the $r$ non-zero elements, which is inspired from \cite{deng2018l0}. The detailed process is shown in Algorithm \ref{RPP}. The initial candidates are the largest $r+r_0$ elements, instead of $r$. Then the RPP iteratively selects elements from the candidates through probabilistic sampling until the $L_0$ norm of $\pmb{\beta}^j_{L_0}$ reaches $r$. The sampling probability is proportional to the absolute value of $\beta^{ij}$ (i.e. $\beta_{abs}^{ij}$). At each iteration, the selected elements will be removed from the candidate set.

\begin{figure}[!htbp]
\centering
\includegraphics[width=0.48\textwidth]{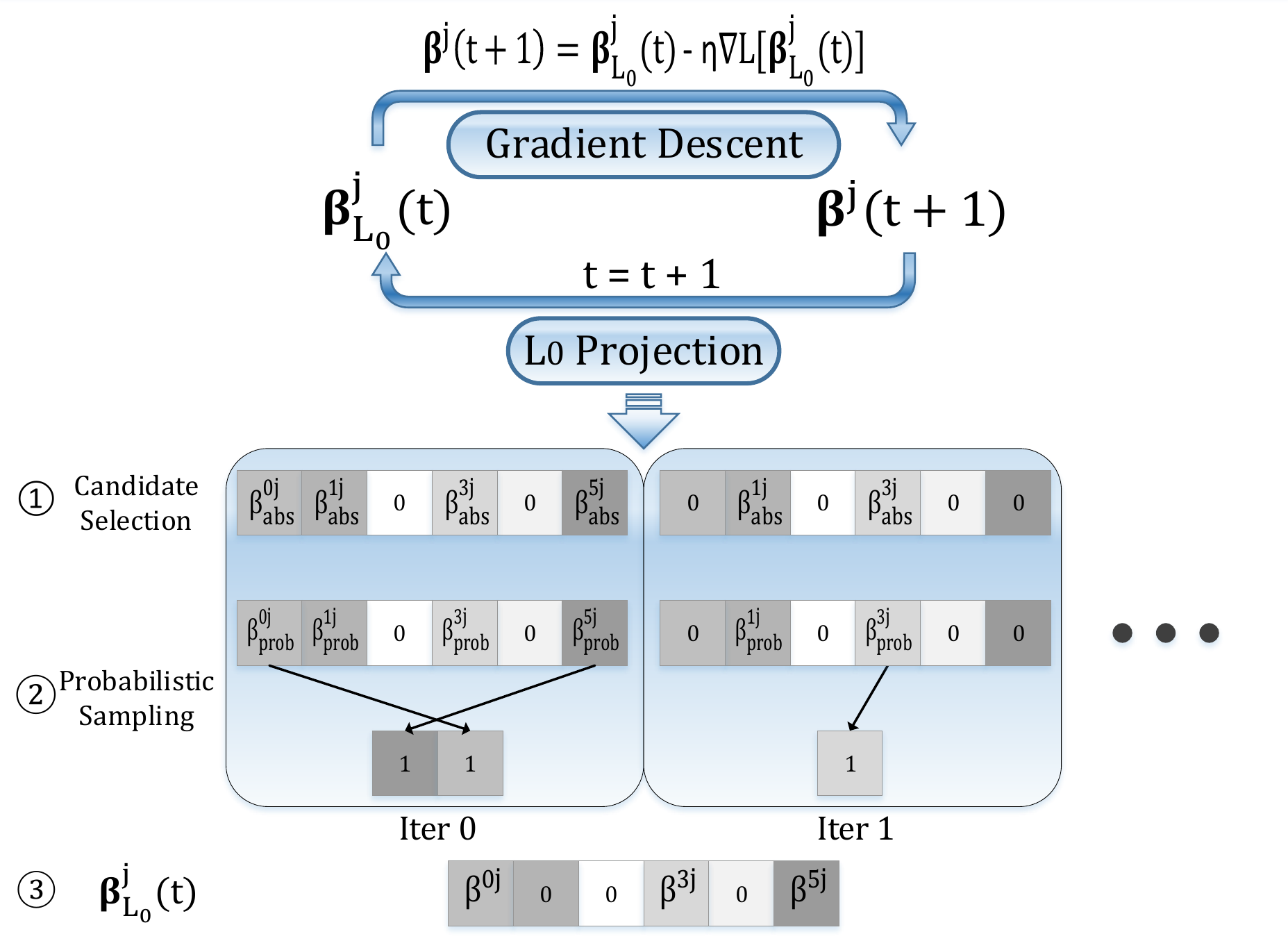}
\caption{Illustration of LGD and RPP.}
\label{L0 Projection}
\end{figure}

\begin{algorithm}
\label{LGD}
\caption{$L_0$ norm constrained gradient descent (LGD)}
\KwData{$\pmb{X}$, $\pmb{Y}$, $L_0$ norm $r$, relaxation factor $r_0$, learning rate $\eta$}
\KwResult{Binary pruning mask $\pmb\beta^j_{L_0}$ with $r$ ones}
\BlankLine
\BlankLine
$\pmb{\beta}^j_{L_0}=RPP(randn(0,1))$;\\
\While{not converge}{
	// update $\pmb{\beta}^j$ through gradient descent in equation (\ref{GD})\\
	$\pmb{\beta}^j=\pmb{\beta}^j_{L_0}-\eta(\pmb{X}^T\pmb{X}\pmb{\beta}^j_{L_0}-\pmb{X}^T\pmb{Y})$;\\
    \BlankLine
    // transform $\pmb{\beta}^j$ to $\pmb{\beta}^j_{L_0}$ with $L_0$-norm constraint\\
    $\pmb{\beta}^j_{L_0}=RPP(\pmb{\beta}^j,~r,~r_0)$
    
    \BlankLine
    // optimize $\pmb{\beta}^j_{L_0}$ through linear regression\\
    $\pmb{Z}=\pmb{X}\pmb\beta^j_{L_0}$\\
    $\alpha=(\pmb{Z}^T\pmb{Z})^{-1}\pmb{Z}^T\pmb{Y}$\\
    $\pmb\beta^j_{L_0}=\alpha \pmb\beta^j_{L_0}$
}
\For{all $i$}{
	$\beta^{ij}_{L_0} \leftarrow 1, \ if \ \beta^{ij}_{L_0} \neq 0$\\
}
\BlankLine
\end{algorithm}

Based on the RPP, now we explain the LGD for solving the $L_0$-norm constrained optimization problem in Equation (\ref{L0 norm}). The overall idea is to integrate the RPP into the normal gradient descent at every iteration. Algorithm \ref{LGD} shows the whole process. The gradient descent is governed by 
\begin{equation} \label{GD}
\begin{aligned}
	\pmb{\beta}^j& = \pmb{\beta}_{L_0}^j - 
    \eta\nabla_{{\beta}_{L_0}^j}\|\pmb{Y}-\pmb{X\beta}_{L_0}^j\|_2^2\\
    & =\pmb{\beta}^j_{L_0}-\eta(\pmb{X}^T\pmb{X}\pmb{\beta}^j_{L_0}-\pmb{X}^T\pmb{Y}),
\end{aligned}
\end{equation}
which is a modified version that frequently switches between the full space of $\pmb{\beta}^j$ and $L_0$-norm constrained space of $\pmb{\beta}^j_{L_0}$. The space switching is implemented by the aforementioned RPP. Note that at each LGD iteration, $\pmb{\beta}_{L_0}^j$ is scaled by a factor $\alpha$ to minimize the loss function in Equation (\ref{L0 norm}) through linear regression. In this work, the number of iterations for LGD is set to 50. At the end, we binarize all the elements in $\pmb{\beta}_{L_0}^j$, which generates the final binary pruning mask. $\pmb{\beta}_{L_0}^{ij}=0$ means the crossbar (crossbar-grain) or the crossbar columns (column-grain) connecting the $i$-th input FM group and $j$-th output FM group can be removed.

Figure \ref{L0 Projection} illustrates the LGD work flow and shows an RPP example. It starts with a randomly initialized $\pmb{\beta}^{j}(0)$. In each outer loop, the pruning coefficient vector $\pmb{\beta}_j$ is firstly calculated through gradient descent, and then the $L_0$-norm constrained vector $\pmb{\beta}_{L_0}^{j}$ is updated through RPP. The inner loop demonstrates how RPP works. Suppose the length of coefficient vector $\pmb{\beta}^j$ is 6, $r=3$, and $r_0=1$. The rectangle with darker shade indicates larger absolute value. Since the absolute value of $\beta^{2j}$ and $\beta^{4j}$ are smaller than others, they are screened out of candidate set at the beginning, i.e. $\beta_{abs}^{2j}=\beta_{abs}^{4j}=0$. The rest four elements ($r+r_0=4$) forms the candidate set. Through the probabilistic sampling strategy in RPP, $\beta_{abs}^{0j}$ and $\beta_{abs}^{5j}$ are sampled and removed from the candidate set at RPP iteration 0; furthermore, $\beta_{abs}^{3j}$ is selected from the candidate set of $\{\beta_{abs}^{1j},~\beta_{abs}^{3j}\}$ at the next iteration. Because the $L_0$ norm of $\pmb{\beta}_{L_0}^{j}(0)$ has reached 3, the RPP ends after two iterations in this example. It is interesting that, although we have $\beta_{prob}^{1j} \geqslant \beta_{prob}^{3j}$, $\beta_{prob}^{3j}$ is finally selected. This indicates that the RPP gives the sub-largest elements some opportunities, and provides better convergence. 

In a nutshell, the proposed LGD along with RPP has several advantages: 1) it is able to convert the NP-hard $L_0$-norm constrained optimization problem to be approximately solvable; 2) it can accurately control the final sparsity by tuning the value of $r$; 3) the smart relaxation factor $r_0$ can provide better convergence by introducing probabilistic projection.

\subsection{Input FMs Reorder}\label{sec:permutation}
The crossbar-aware pruning considers the grouping effect of crossbars. The above analysis does not consider how to group the input or output FMs. Recall that in Figure \ref{pruning}, the order of output FMs has no influence on the final results because they are independent. However, the order of input FMs matters because all the crossbar columns share the same input rows. In above sections, we use the simplest original order when mapping the input FMs. In this section, we design a reorder strategy to tune the input FM order and reduce the pruning error. 

The reorder of input FMs tries to increase the possibility that the more important input FMs can be clustered into the same group. In this way, the impact of pruning other less important groups on model accuracy can be reduced. Usually, a larger partial sum $|\pmb{Y}_S^{p,q}|$ indicates more contribution to final complete sum of output FM. The importance of each input FM is identified by summing all $\pmb{Y}_S^{p,q}$ with the same $p$, which is calculated by
\begin{equation} \label{order}
	\pmb{Y}_F^p = \sum\nolimits_{q} \pmb{Y}_S^{p,q}.
\end{equation}
Next, we reorder the input FMs according to the importance values. The following pruning process is the same as that without reorder. Note that we do not take the absolute value of $\pmb{Y}_S^{p,q}$, since the crossbar output tends to be zero, i.e. little contribution, if the largest positive $\pmb{Y}_S^{p,q}$ and the smallest negative $\pmb{Y}_S^{p,q}$ (with similar absolute value) fall into the same crossbar. This will lead to less distinguishable between the large partial sums and small partial sums.

The input FMs reorder works well for the front layers, however the deeper layers cannot benefit much from this technique. This might be because that in deep layers, each input FM has extracted high-level abstraction and has `equal' importance. We provide a detailed discussion in Section \ref{sec:Results}.

\subsection{Crossbar-aware Single-layer Level Pruning}
Our pruning framework runs based on a pre-trained model. For the single-layer level pruning in this section, other layers keep unchanged except the layer to be pruned. Before implementing the pruning methodology presented in Algorithm \ref{alg prune}, the order of input FM groups should be fixed (with or without reorder). The layer pruning is conducted group by group along the output FM dimension (from $j=0$ to $j=J-1$). Suppose $N$ images are sampled for pruning. $\pmb{FM}_{Y_{C1}^j} \in R^{N \times S_{out} \times I \times K_{out}}$ and $\pmb{FM}_{Y_{C2}^j} \in R^{N \times S_{out} \times K_{out}}$  are the concatenation tensor representation of the partial sum and complete sum, respectively, involving the $j$-th output FM group. For each output FM group, we firstly reshape $\pmb{FM}_{Y_{C1}^j}$ and $\pmb{FM}_{Y_{C2}^j}$ to a matrix and vector format, respectively. Then, we run the LGD to generate the binary pruning mask $\pmb{\beta}_{L_0}^j$ for pruning the crossbars (crossbar-grain) or crossbar columns (column-grain). In the latter one, we need to recombine the non-zero crossbar columns to assemble new crossbars, like that in Figure \ref{pruning}. Note that as many pruning works do,  a final linear regression is required to tune the weights \cite{luo2017thinet, he2017channel}. In particular, according to the pruning mask, we can collect the required input and output data for linear regression as $\pmb{LR}_X$ and $\pmb{LR}_Y$, respectively. Then, the remained weights involving the $j$-th output FM group are tuned by
\begin{equation} \label{LR}
	\pmb{W}_{rem}^j = LR(\pmb{LR}_X,\pmb{LR}_Y)=(\pmb{LR}_X^T\pmb{LR}_X)^{-1}\pmb{LR}_X^T\pmb{LR}_Y.
\end{equation}
Note that $k^2$ in Algorithm \ref{alg prune} is the size of each weight kernel, e.g. 3$\times$3, and we have $\pmb{FM}_{L_x} \in R^{N \times S_{out} \times k^2 \times P}$ and $\pmb{FM}_{L_y} \in R^{N \times S_{out} \times Q}$.

%alg3 whole structure
\begin{algorithm}
\label{alg prune}
\caption{Crossbar-aware pruning}
\KwData{$\pmb{FM}_{Y_{C1}^j}$, $\pmb{FM}_{Y_{C2}^j}$, $\pmb{FM}_{L_X}$, $\pmb{FM}_{L_Y}$, $L_0$ norm $r$, relaxation factor $r_0$, learning rate $\eta$}
\KwResult{Weight update}

\For{j=0:J$-$1}{
	\BlankLine
    // reshape $\pmb{FM}_{Y_{C1}^j}$ to a 2-D matrix for pruning\\
    $\pmb{prune}_X=reshape(\pmb{FM}_{Y_{C1}^j},~(N S_{out} K_{out}, ~I))$;\\
    \BlankLine
    // reshape $\pmb{FM}_{Y_{C2}^j}$ to a 1-D vector for pruning\\
    $\pmb{prune}_Y=reshape(\pmb{FM}_{Y_{C2}^j},~N S_{out} K_{out})$;\\
    \BlankLine
    \BlankLine
    // execute $L_0$-norm gradient descent (LGD) algorithm\\   $\pmb{\beta}^j_{L_0}=LGD(\pmb{prune}_X,~\pmb{prune}_Y,~r,~r_0,~\eta)$;\\
    \BlankLine
    \BlankLine
    // collect remained input FMs\\
    $G_{in}=\{g_{in}^i\ |\ \beta^{ij}=1,~i=0, 1,.., I-1\}$;\\
    \BlankLine
    \BlankLine
    // linear regression to minimize pruning loss\\
    $\pmb{LR}_X=reshape(\pmb{FM}_{L_X}[:, :, :, G_{in}],$\\$~~~~~~~~~~~~~~~~~~~~(N S_{out}, ~k^2 length(G_{in}))$;\\
    $\pmb{LR}_Y=reshape(\pmb{FM}_{L_Y}[:, :, g_{out}^j],~(N S_{out}, ~K_{out}))$;\\
    $\pmb{W}_{rem}^j\leftarrow LR(\pmb{LR}_X,\pmb{LR}_Y)$; // linear regression
  }
\end{algorithm}

\begin{figure}[!htbp]
\centering
\includegraphics[width=0.49\textwidth]{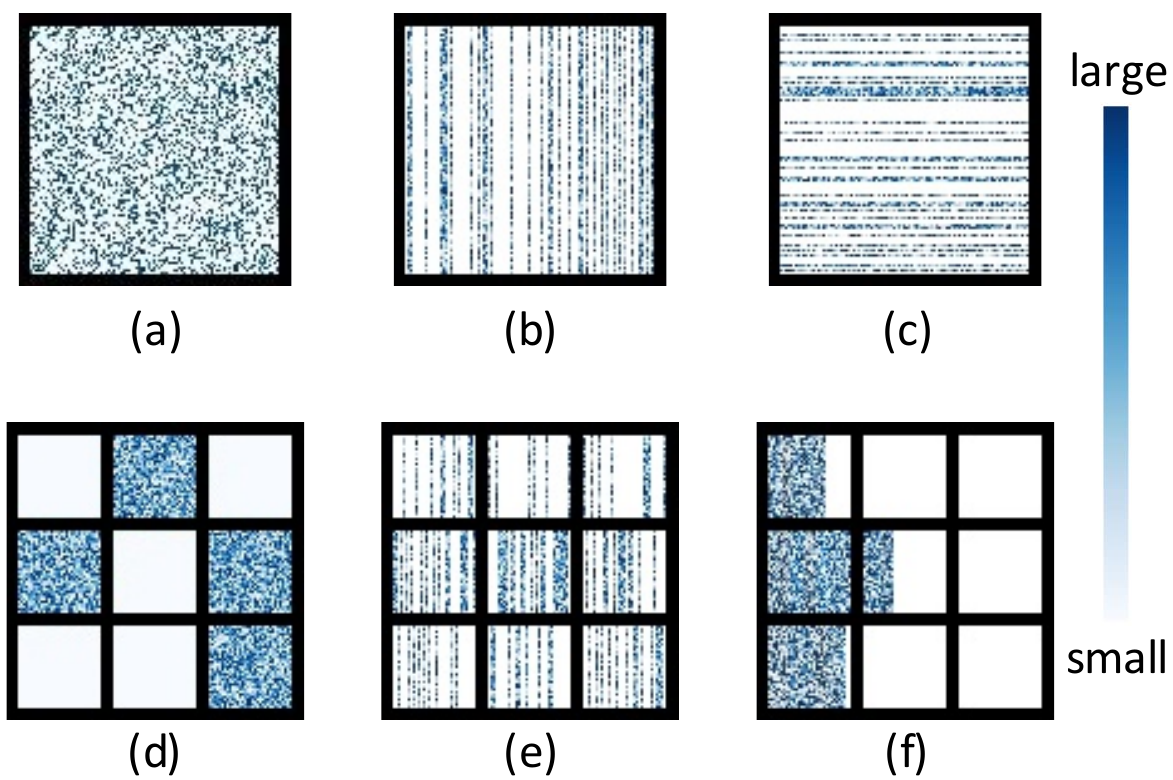}
\caption{Comparison of sparse pattern resulted from different pruning strategies: (a) element-wise pruning \cite{han2015learning, han2015deep}. (b) vector-wise (row) pruning \cite{wen2016learning, wen2017learning}. (c) vector-wise (column) pruning \cite{wen2016learning, wen2017learning}. (d) crossbar-grain pruning. (e) column-grain pruning before recombination. (f) column-grain pruning after recombination. The grid in (d)-(f) denotes crossbars.}
\label{pruning pattern}
\end{figure}

For clarity, Figure \ref{pruning pattern} compares the sparse pattern resulted from five different pruning strategies. In Figure \ref{pruning pattern}(a)-(c), the matrix corresponds to the GEMM weight matrix after transforming the Conv operation into matrix-matrix multiplication \cite{cho2017mec, ding2017c}, rather than the crossbar grids in Figure \ref{pruning pattern}(d)-(f). Although the element-wise pruning \cite{han2015learning, han2015deep} and vector-wise pruning \cite{wen2016learning, wen2017learning} can produce unstructured and structured sparsity, respectively, they only consider the original logic computational graph or the execution graph on general purpose platform (e.g. CPU/GPU). Therefore, they cannot be directly used by crossbar devices. The crossbar-grain pruning is straightforward to be used by the crossbar architecture, whereas, it suffers from larger accuracy degradation. The column-grain pruning produces sparsity in crossbar columns (not the matrix columns in the vector-wise pruning) which improves the accuracy with smaller pruning grain. By recombining the remained non-zero crossbar columns, column-grain pruning is able to remove many redundant crossbars. 

Note that the FC layer can also adopt our pruning framework to reduce crossbar overhead. FC layer corresponds to 1$\times$1 FM size in Conv layer. Since the number of output neurons in FC layer is large (e.g. 2048), we set $K_{out}=8$ in this case instead of employing the column-grain sparsity for Conv layer directly. The recombination of the non-zero columns is still used in FC layer pruning.  

\subsection{Crossbar-aware Network Level Pruning}\label{sec:network pruning}
A key principle in neural network pruning is to balance the model sparsity and accuracy loss \cite{li2016pruning}. Distinguishing from previous work, we consider the tradeoff between crossbar overhead and accuracy loss. Since the tolerance to sparsity for each layer is variable, different layers should adopt different pruning ratios (sparsity of $\beta$) to minimize redundant crossbars as much as possible, while maintaining the model accuracy. 

In network pruning we will first try single-layer pruning (each time only one layer is pruned) under different pruning ratios (i.e, from 20\% to 70\% with 5\% interval) based on Algorithm \ref{alg prune}. Then each layer{'}s tolerance to sparsity can be reflected from the accuracy drop curve. Thus, we determine how many crossbars can be pruned according to a pre-defined accuracy drop threshold $T_d^{initial}$. Specifically, we choose the pruning ratio that has the closest accuracy drop to $T_d^{initial}$ (but larger than $T_d^{initial}$) for each layer. 

Based on the initial pruning ratio, we further design three conditions to finally stop the single-layer pruning and determine the pruning ratio for every layer: 1) when the accuracy drop is larger than a threshold $T_d$; 2) when the pruning ratio is larger than a threshold $T_p$; 3) when the number of remained crossbars are smaller than a threshold $T_c$. Once any of them is satisfied, the single-layer pruning stops and the pruning ratio is finally fixed for that layer. The first condition is designed to maintain the final accuracy after pruning. The later two conditions try to control the resource overhead for each layer. There is no need to continue pruning if the pruning ratio in that layer is already small. In the large network like VGG16, the crossbar overhead presents a large variance among different layers due to different layer size. Therefore, the pruning ratio is not enough to accurately reflect the crossbar overhead. To this end, we introduce an additional threshold $T_c$ to jointly control the resource consumption. A detailed example can be found in Figure \ref{threshold}. Then, we prune the whole network layer by layer according to the finally determined pruning ratios of all layers. A fine-tuning step is also required to restore the network accuracy at the end as previous work \cite{han2015learning, han2015deep}.

\section{Experimental Results}\label{sec:Results}
\subsection{Experiment Setup}

\textbf{CNN Model}. We validate our experiments on CIFAR-10 \cite{krizhevsky2009learning} and ImageNet \cite{deng2009imagenet} datasets. We use three CNN models to run our crossbar-aware pruning. On CIFAR-10 dataset, an eight-layer network (VGG8 \cite{li2016ternary, deng2018gxnor}) is adopted, and on ImageNet dataset, VGG16 \cite{simonyan2014very} and ResNet18 \cite{he2016deep} are used.

\begin{figure}[!htbp]
\centering
\includegraphics[width=0.48\textwidth]{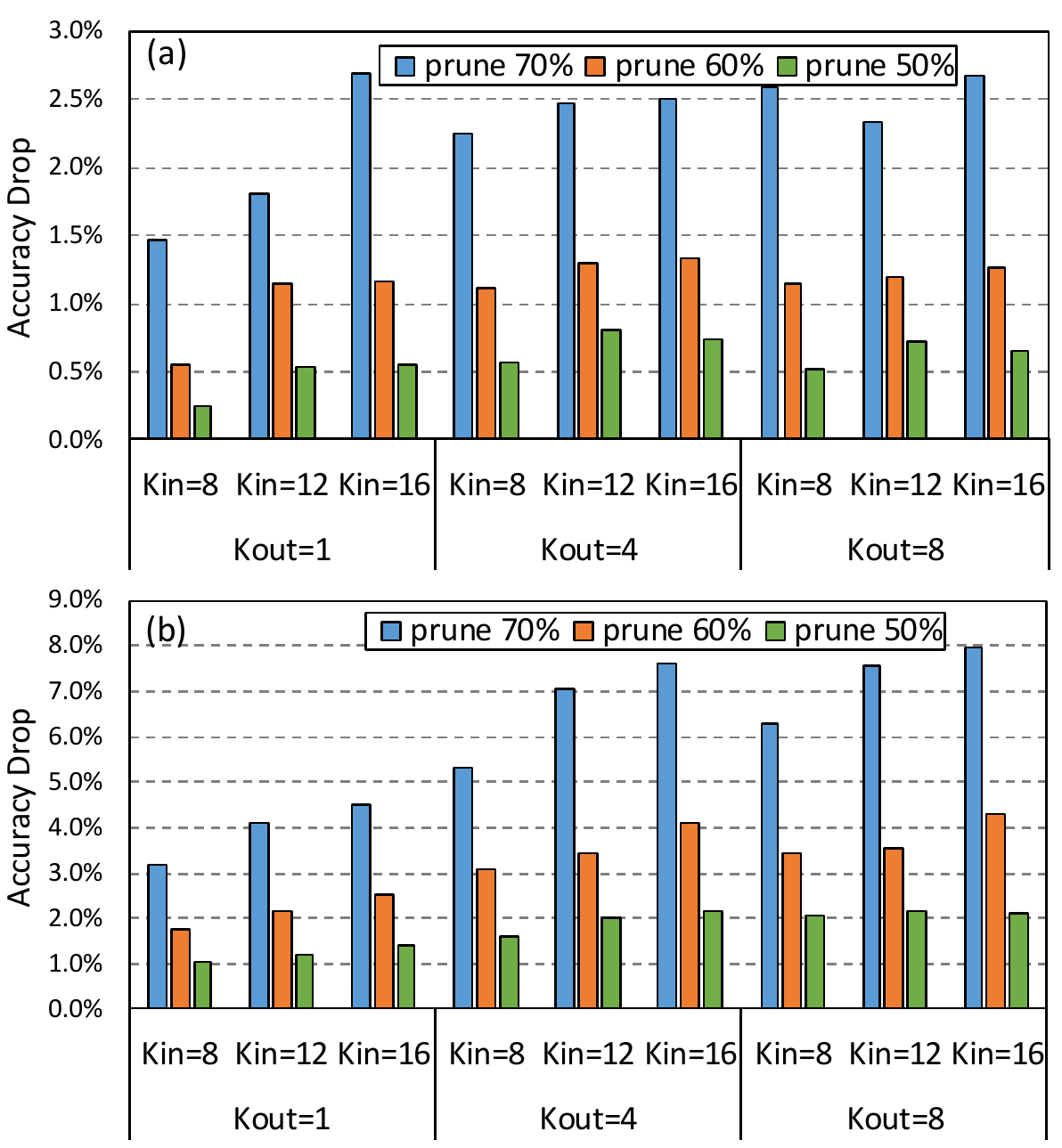}
\caption{Accuracy drop with different group size of output FMs: (a) Conv2-2 (VGG16), (b) Conv4-2 (VGG16).}
\label{cin cout}
\end{figure}

\textbf{Crossbar Platform}. For one-to-one comparison, we adopt the same semi-folded mapping compiler for the neural network chip in \cite{deng2018semi}. To reduce the development period and save fabrication cost, we use the off-the-shelf SRAM array with extra multipliers and accumulators to simulate the crossbar-based VMM engine, like that in \cite{merolla2014million} and \cite{7409624}. Please note that our crossbar-aware pruning is a general framework for any crossbar architecture. 

\begin{figure*}[!htbp]
\centering
\includegraphics[width=0.91\textwidth]{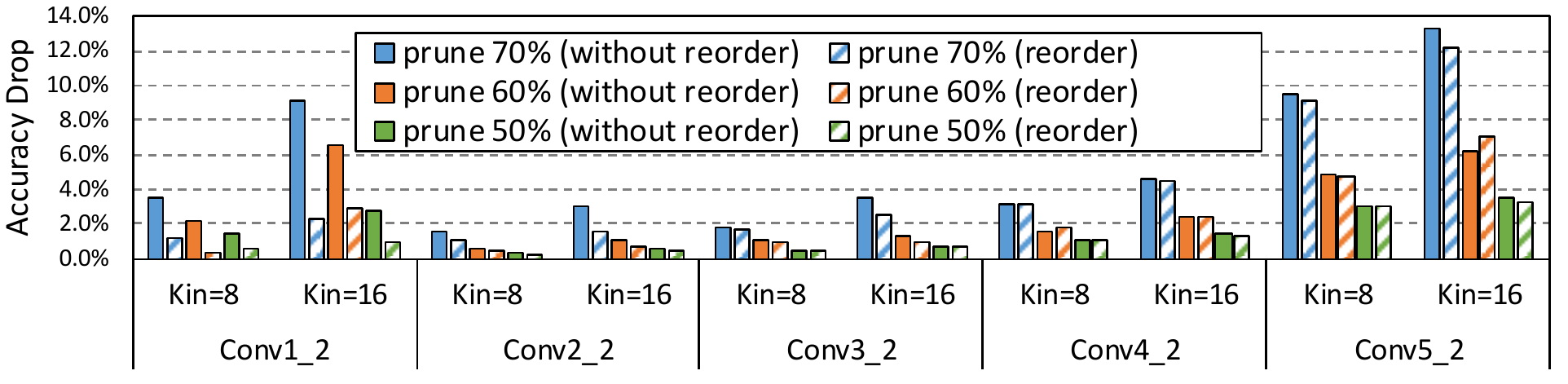}
\caption{Accuracy drop under original index order (without reorder) and descending importance order (reorder) of input FMs for different layers in VGG16.}
\label{permute}
\end{figure*}

\textbf{Pruning Configuration}. In the following experiments, 5000 images are sampled from the training dataset, and each class have the same number of images. For each image, 10 points are sampled from each FM (size of $S_{out}$) for LGD, and we change 10 to 2 for the final linear regression. 

\subsection{Analysis of Single-layer Level Pruning}
In this section we will use the pruning result for single layer to analyze the influence of $K_{in}$, $K_{out}$, pruning ratio, and the input FM order. The analysis conclusions in this section can guide the next network pruning. Because the accuracy gap among different configurations could be narrowed down via fine-tuning technique, which will impede the sensitivity study, we abandon the final fine-tuning step in this section.

Figure \ref{cin cout} shows the accuracy drop for two layers in VGG16 (Conv2-2 and Conv4-2) under different $K_{in}$, $K_{out}$ settings and pruning ratios (input FMs are grouped with reorder). As previous analysis in Section \ref{sec:sparsity grain}, decreasing $K_{out}$ can mitigate the dependency requirement among each $Y_C^{i,q}$ in the same $j$-th output FM group and then reduce the pruning error. Apparently, the smallest accuracy drop in both Conv2\_2 and Conv4\_2 are produced from the column-grain sparsity ($K_{out}=1$). This is consistent with our prediction in Section \ref{sec:sparsity grain} that column-grain sparsity can achieve higher accuracy than crossbar-grain sparsity ($K_{out}>1$). For the smaller pruning ratio (50\% and 60\%), the accuracy difference between $K_{out}=4$ and $K_{out}=8$ is not obvious. In contrast, for large pruning ratio (70\%), $K_{out}=8$ overally performs worse than $K_{out}=4$. This indicates the influence of $K_{out}$ will be amplified in the case of large pruning ratio. Generally speaking, smaller $K_{out}$ often brings better accuracy. Similar to $K_{out}$, $K_{in}$ also impacts the model accuracy. Larger $K_{in}$ usually suffers from larger accuracy drop due to the coarser pruning grain. However, different $K_{in}$ brings very different resource overhead in the semi-folded mapping framework \cite{deng2018semi} and it is deeply coupled with the mapping implementation. To simplify the problem and focus on our pruning framework, we initially determine $K_{in}$ for each layer towards the minimum crossbar overhead when we mapping the dense model onto crossbar, and fix them during the next pruning process. The $K_{in}$ configuration details can refer to previous work \cite{deng2018semi}.

Since the column-grain pruning can obviously outperform the crossbar-grain pruning from the perspective of model accuracy with the same sparsity, we will focus on the column-grain pruning in the following experiments. In other words, we will set $K_{out}=1$.

\begin{figure}[!htbp]
\centering
\includegraphics[width=0.47\textwidth]{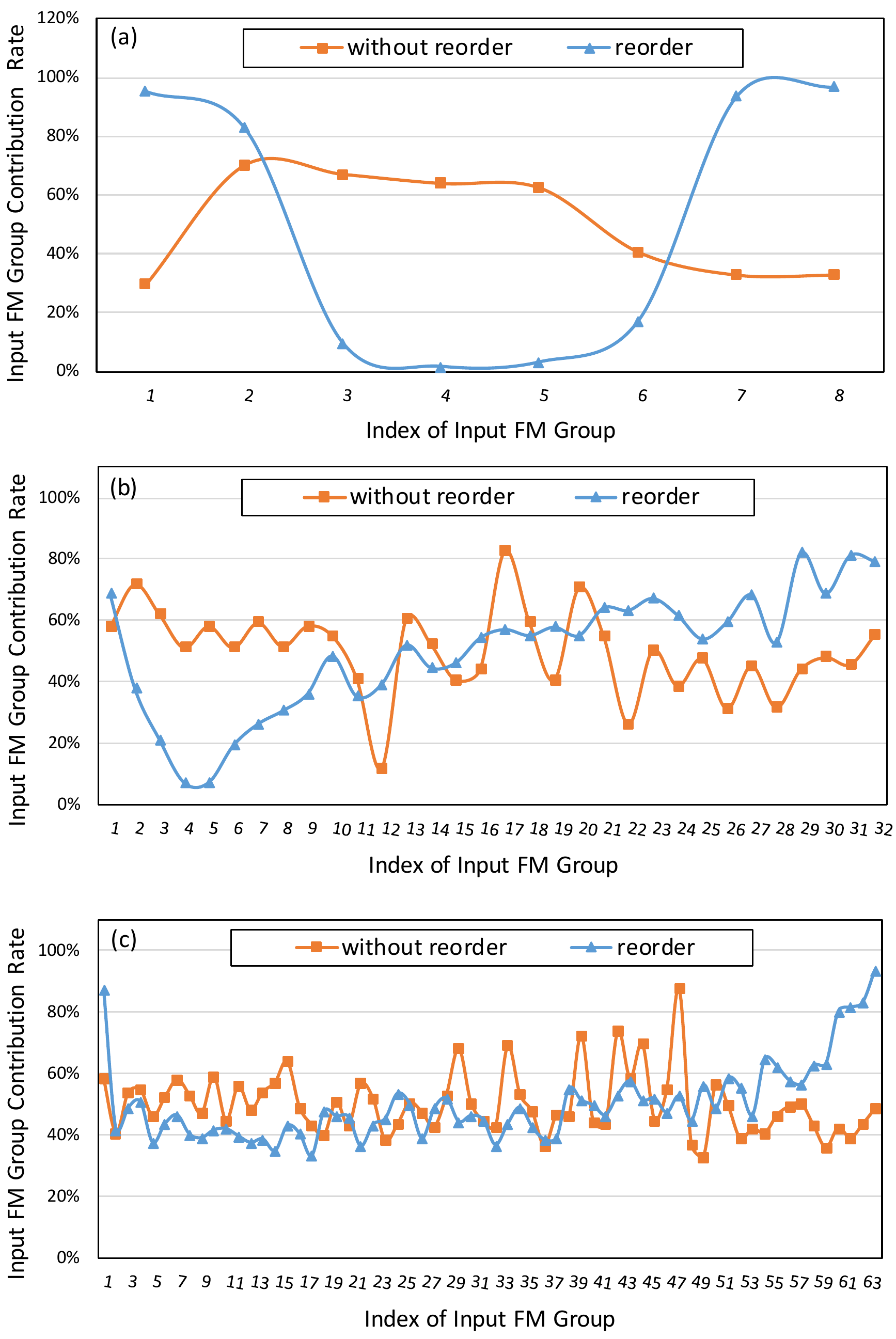}
\caption{Contribution rate of input FM groups under original index order (without reorder) and descending importance order (reorder) for different layers in VGG16: (a) Conv1\_2, (b) Conv3\_2, (c) Conv5\_2.}
\label{beta}
\end{figure}

Next we study the influence of the reorder of input FMs. We select five layers in VGG16 to present the accuracy change by adding the reorder operation. The index reorder is according to the descending order of the importance of input FMs calculated by Equation (\ref{order}). As shown in Figure \ref{permute}, input FMs reorder significantly improves the accuracy in the front layers, especially for large pruning ratio (70\%). While in deep layers, the improvement is degraded. As mentioned in Section \ref{sec:permutation}, the reason might be that in deep layers, each input FM has extracted high-level abstraction and has `equal' importance.

In order to clearly visualize how input FMs reorder works, Figure \ref{beta} shows the contribution rate of input FM groups from three difference layers in VGG16 with or without input FMs reorder. Here the contribution rate of input FM group indicates the percentage of output FM groups that  uses this input FM group (i.e. it has contribution). It can be calculated by $(\sum_j \beta_{L_0}^{ij})/J$. The global pruning rate in this figure is fixed at 50\%. For the original index order, the contribution rate of input FM groups shows a fluctuated evenly distribution, which means no input FM group performs absolutely important or unimportant. This will cause larger pruning error. In stark contrast, after using the reorder strategy, we can cluster the trivial and non-trivial FMs into different FM groups. This phenomenon is quite clear in the front layers. For example, in Conv1\_2, the contribution rates in middle input FM groups are much lower than those in head and tail, which indicates we successfully clustered the non-trivial input FMs into the two terminals and pruning the middle trivial input FMs can still maintain the model accuracy well. However, it should be noted that as the layer goes deeper, most of input FM groups tend to be equally important even if the ones in head and tail still have a little higher contribution rate. The reason has been explained earlier that each FM extracts high-level abstraction in deep layers so that all of them probably have necessary contribution to the final recognition.  

\subsection{Analysis of Network Level Pruning}
In this section we will conduct the pruning of the whole network. In our network pruning the first Conv layer and the last FC layer are not pruned. The reason is that these two layers serve as the first feature extraction layer and the last classification layer, pruning of which renders significant accuracy drop. Additionally, the resource cost in these two layers is negligible, so there is no need to prune them.

\begin{figure}[!htbp]
\centering
\includegraphics[width=0.46\textwidth]{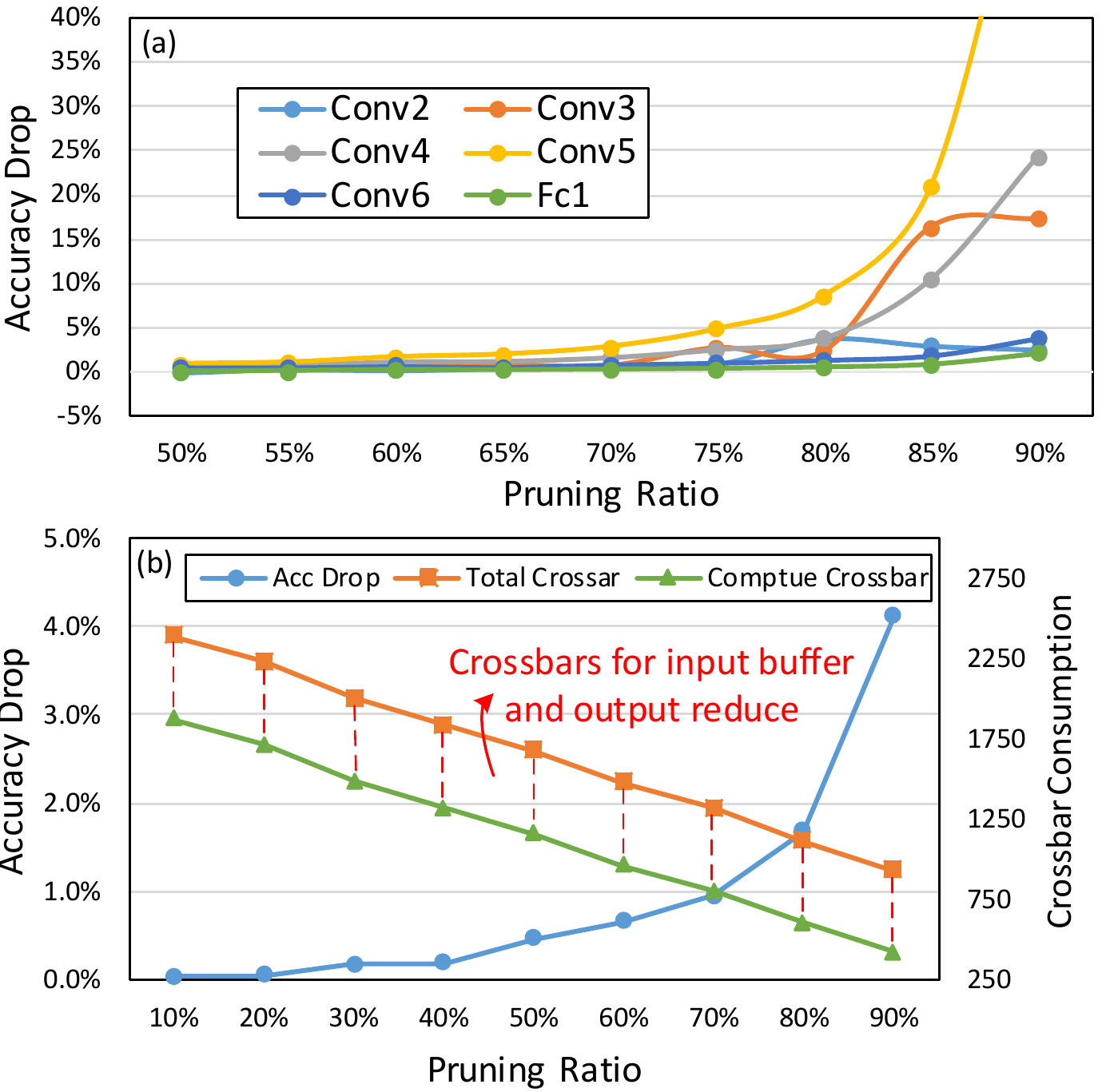}
\caption{Accuracy drop and crossbar overhead in VGG8.}
\label{vgg_single}
\vspace{-5pt}
\end{figure}

Before stepping into our network pruning strategy described in Section \ref{sec:network pruning}, we will first discuss the variance of sparsity tolerance in each layer, as well as the relation between accuracy drop and crossbar overhead. Figure \ref{vgg_single}(a) shows the accuracy drop when separately pruning each layer under different pruning ratio in VGG8. When the pruning ratio is less than 70\%, all layers can maintain the accuracy drop within 3\%. The accuracy drop will significantly increase when the pruning becomes more aggressive. Interestingly, different layers present different tolerance to sparsity. For example, Conv2, Conv6, and FC1 seem more robust compared to other layers. Figure \ref{vgg_single}(b) depicts the accuracy drop and resource overhead when pruning the whole network under different pruning ratio. For simplicity, here we prune all the layers with the same pruning ratio. The accuracy drop is controlled within 0.5\% when the pruning ratio is smaller than 50\%. Similarly, more aggressive pruning will cause significantly increasing accuracy degradation. For the crossbar overhead, nearly linear resource saving is observed as pruning ratio increases. Only the compute crossbars are reduced, and the crossbars for input buffer and output reduce cannot be saved. More details can be found in the original semi-folded mapping paper \cite{deng2018semi}. Note that Figure \ref{vgg_single}(a) doesn't use the fine-tuning technique but \ref{vgg_single}(b) does. The underlying reason is that the single-layer pruning aims to help sensitivity analysis so we expect larger accuracy gap, while the network pruning always wants higher accuracy for practical implementation.

\begin{table}[!htbp]
\caption{Comparison of crossbar overhead and model accuracy in VGG8 before and after pruning. Three different pruning ratios are used. `T': total crossbar, `C': compute crossbar.}
\vspace{5pt}
\label{vgg8 detail}
\centering
\renewcommand\arraystretch{1.3}
\resizebox{0.48\textwidth}{!}{
\begin{tabular}{||c | c c | c c | c c | c c ||}   
   \hline
   Pruning & \multicolumn{2}{c|}{0\% (dense)} & \multicolumn{2}{c|}{10\%} & \multicolumn{2}{c|}{50\%}  & \multicolumn{2}{c||}{90\%}\\
   Ratio   & T & C & T & C & T & C & T & C\\
   \hline
   
   Conv2 & 176 & 128 & 164 & 116 & 124 & 76  & 72  & 32\\
   Conv3 & 224 & 160 & 208 & 152 & 156 & 100 & 88  & 36\\
   Conv4 & 604 & 512 & 570 & 478 & 376 & 284 & 168 & 76\\
   Conv5 & 364 & 304 & 352 & 296 & 222 & 166 & 94  & 38\\
   Conv6 & 698 & 592 & 682 & 576 & 424 & 318 & 182 & 76\\
   Fc1   & 167 & 128 & 161 & 122 & 120 & 81  & 71  & 32\\
   \hline
   
   Accuracy         & \multicolumn{2}{c|}{94.29\%} & \multicolumn{2}{c|}{94.27\%}
   			   & \multicolumn{2}{c|}{93.84\%} & \multicolumn{2}{c||}{90.18\%}\\
   T & \multicolumn{2}{c|}{2484}    & \multicolumn{2}{c|}{2388}
   			   & \multicolumn{2}{c|}{1673}    & \multicolumn{2}{c||}{926}\\
   C & \multicolumn{2}{c|}{1954}    & \multicolumn{2}{c|}{1870}
   			   & \multicolumn{2}{c|}{1155}    & \multicolumn{2}{c||}{420}\\
   \hline

\end{tabular}}
\end{table}

\begin{figure*}[!htbp]
\centering
\includegraphics[width=0.98\textwidth]{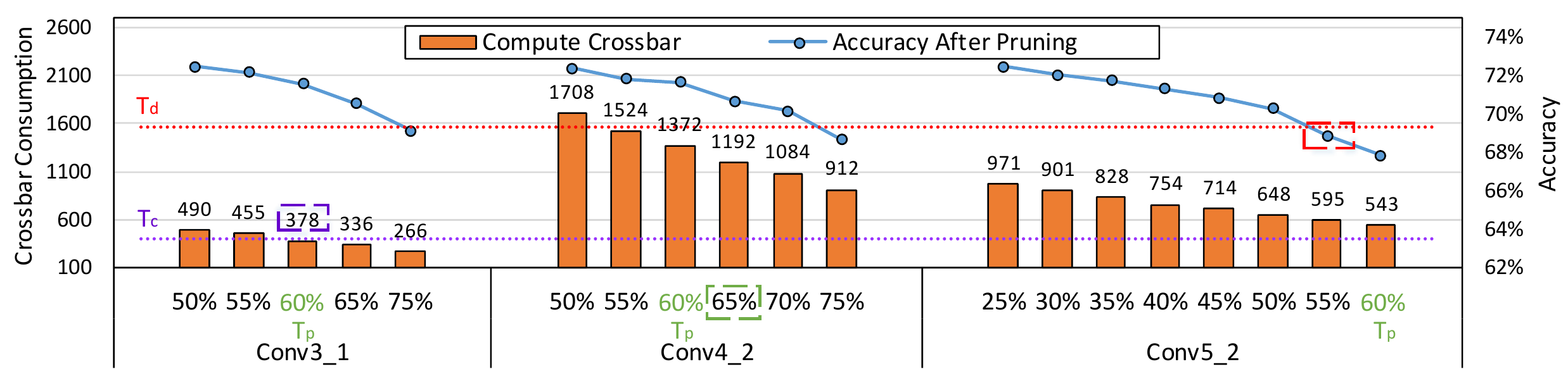}
\caption{Illustration of the three conditions for determining the final pruning ratio of each layer.}
\label{threshold}
\end{figure*}

\begin{figure}[!htbp]
\centering
\includegraphics[width=0.48\textwidth]{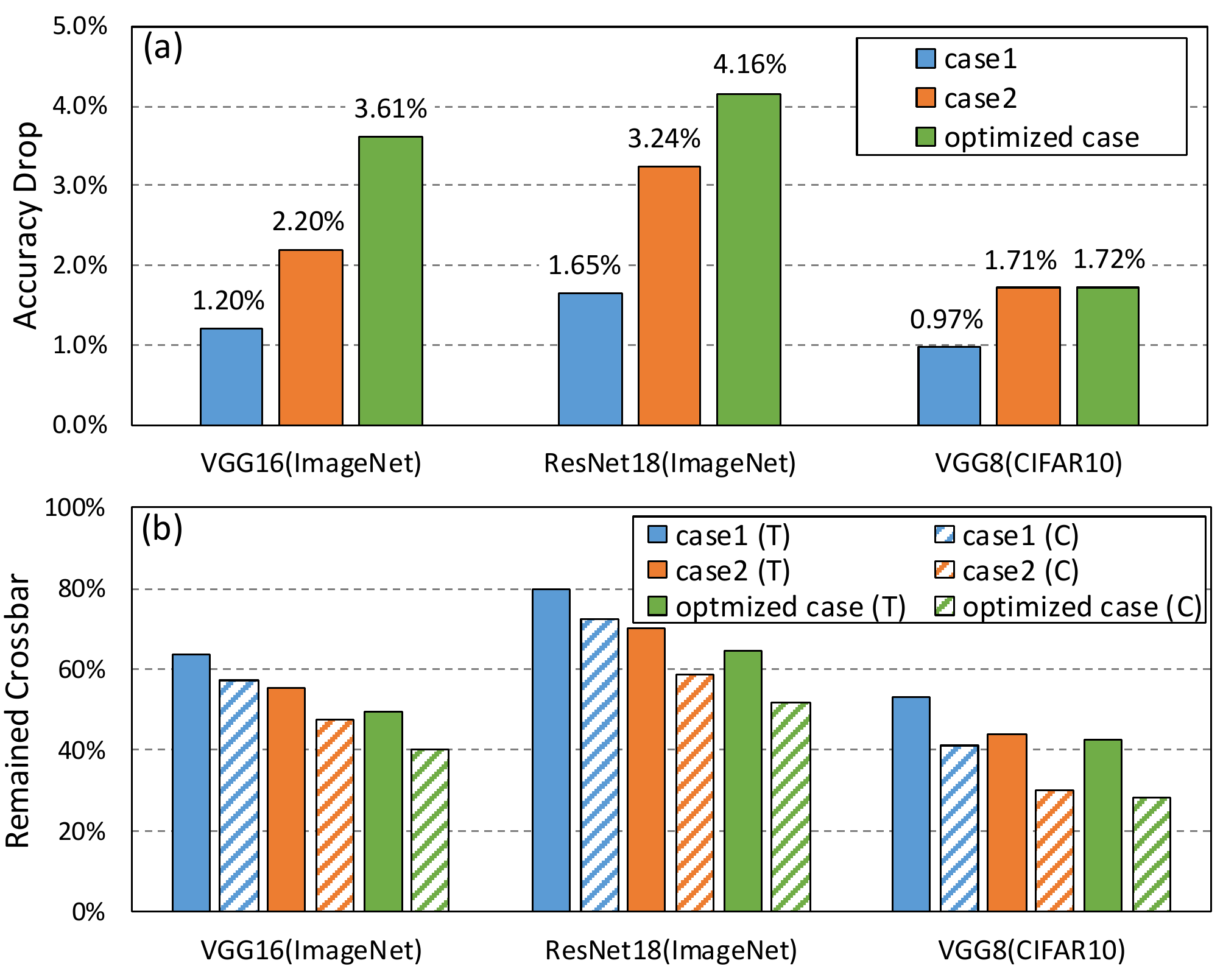}
\caption{Accuracy drop and crossbar overhead in three networks under different pruning configuration. `T': total crossbar; `C': compute crossbar.}
\label{final result}
\end{figure}

\begin{table*}[!htbp]
\caption{Crossbar overhead in three networks under different pruning configuration.}
\vspace{5pt}
\label{final table}
\centering
\renewcommand\arraystretch{1.3}
\resizebox{1\textwidth}{!}{
\begin{tabular}{||c | c c c c | c c c c | c||}   
   \hline
   & \multicolumn{4}{c|}{ \textbf{Total Crossbar}} & \multicolumn{4}{c|}{ \textbf{Compute Crossbar}} &\textbf{Stop Conditions}  \\
   \hline\hline
   
   ImageNet & 0\% (dense) & case1 & case2 & optimized case & 0\% (dense) & case1 & case2 & optimized case & \\
   \hline
   VGG16 & 25801 & 16428 & 14319 & 12749 & 21720&12403 & 10294 & 8724 & ~~$T_d=4\%$,~~ $T_p=60\%$,~ $T_c=400$ \\
   \hline
   ResNet18 & 7300 & 5148 & 5815 & 5107 & 4724 & 3718 & 3017 & 2634 & $T_d=3.5\%$, $T_p=60\%$,~ $T_c=65$ \\
   \hline\hline
   
   CIFAR10 & 0\% (dense) & case1 & case2 & optimized case & 0\% (dense) & case1 & case2 & optimized case & \\
   \hline
   VGG8 & 2484 & 1320 & 1092 & 1056 & 1954 & 802 & 582 & 546 & $T_d=4\%$, ~~$T_p=75\%$~~~~~~~~~~~~~\\
   \hline
\end{tabular}}
\end{table*}

In addition to maintain the model accuracy as high as possible after pruning, we also want to maximize the resource reduction. In Table \ref{vgg8 detail}, we select three pruning ratios to show the crossbar overhead in each layer. Conv6 can reduce 516 compute crossbars if the pruning ratio reaches 90\%, however, under the same pruning ratio Conv3 can only save 124 compute crossbars. From both accuracy drop in Figure \ref{vgg_single}(a) and resource saving perspectives, network pruning can benefit more from layer Conv6. Instead of pruning each layer with the same pruning ratio, a higher pruning ratio in Conv6 and a lower pruning ratio in Conv3 may achieve the same accuracy after pruning but reduce more resources.

Motivated by above analysis, we introduce three conditions to determine the pruning ratio for each layer. Based on the conditions of $T_d$ (accuracy drop), $T_p$ (pruning ratio), and $T_c$ (crossbar overhead) in Section \ref{sec:network pruning}, Figure \ref{threshold} shows three layer examples in VGG16 for explaining how it works. Here we set $T_d=4\%$, $T_p=60\%$ and $T_c=400$. For Conv3\_1 layer, the final determined pruning ratio is 60\%. Although it doesn't exceed the accuracy drop threshold $T_d$ and the pruning ratio threshold $T_p$, the crossbar overhead threshold $T_c=400$ has been reached. Continuing to prune cannot gain more significant resource reduction. Similarly, Conv4\_2 selects pruning ratio of 65\% ($>T_p$ condition) and Conv5\_2 selects 55\%  ($>T_d$ condition).

Figure \ref{final result} and Table \ref{final table} show the final results of the whole network pruning with optimized pruning configuration. Here `case1' means the determined pruning ratio for each layer causes $T_d^{initial}=1\%$ accuracy drop when conducting single-layer pruning. Similarly, `case2' corresponds to $T_d^{initial}=2\%$ in VGG16 and ResNet18 and $T_d^{initial}=3\%$ in VGG8. The `optimized case' is the final pruning configuration determined by the three stop conditions for accuracy-resource balance. In VGG8 (CIFAR10), since the network size is much smaller than those on ImageNet, there is no need to design the crossbar threshold $T_c$. 30 extra epochs are used to fine-tune the models on ImageNet (learning rate $10^{-5}$ for the first 20 epochs and $10^{-6}$ for the last 10 epochs), and 60 epochs are used for VGG8 on CIFAR10 (learning rate $10^{-2}$ for the first 30 epochs and $10^{-3}$ for the last 30 epochs). In general, the optimized case can reduce more resources, but at the cost of more accuracy degradation. Finally, we can save 59.8\%, 44.2\%, and 72.1\% compute crossbars in VGG16, ResNet18, and VGG8. Furthermore, we find that the ResNet18 is more sensitive against pruning than the VGG networks due to its compact structure.

\section{Conclusion and Discussion}\label{sec:Conclusion}
In this work we propose a crossbar-aware pruning framework to introduce usable sparsity in CNNs for crossbar-based architecture. We formulate the issue as a constrained optimization problem and design an $L_0$-norm constrained gradient descent (LGD) algorithm to solve it. In LGD, a relaxant probabilistic projection (RPP) is heuristically leveraged to switch between the full space and the sparse space for producing the pruning mask. The LGD can accurately control the sparsity and bring better convergence. In this way, we are able to achieve two pruning grains: crossbar grain and column grain. The former one is straightforward and easy to use, while the latter one can obtain better accuracy and the same level of sparsity through a recombination of the non-zero crossbar columns. Furthermore, a reorder strategy of the input FMs is utilized to reduce the pruning error by clustering the FMs with similar importance. Based on an optimized pruning configuration with three elaborate pruning conditions, finally, 44\%-72\% crossbars can be saved on three benchmarking networks on CIFAR10 and ImageNet. This work provides a new co-design solution for mapping CNNs onto various crossbar devices with significantly less resource overhead and the resulting energy consumption.

\begin{figure}[!htbp]
\centering
\includegraphics[width=0.45\textwidth]{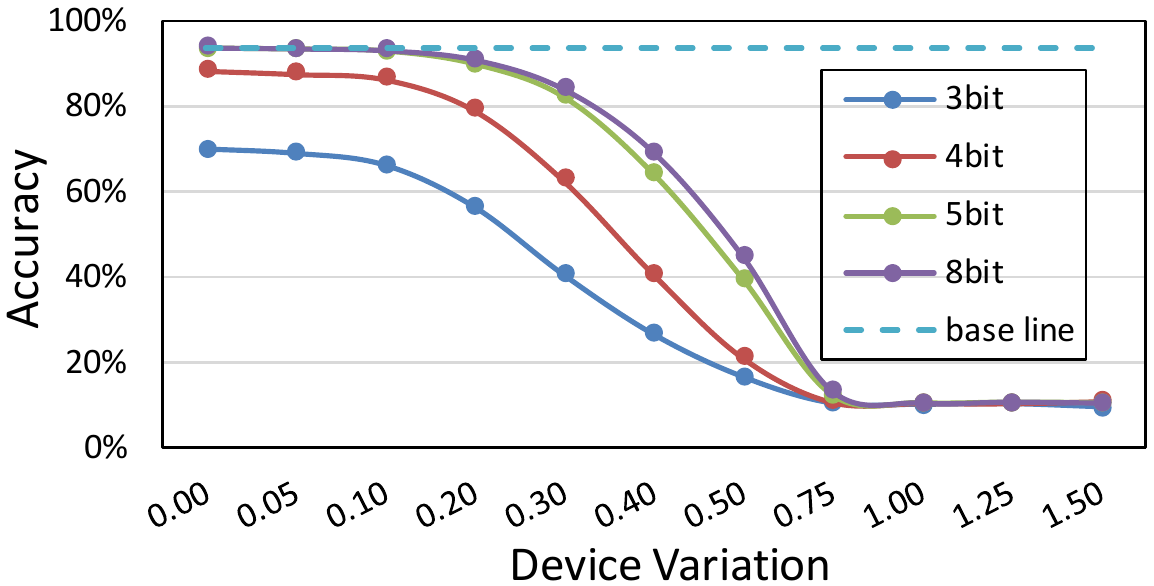}
\caption{Influence of device quantization and variation.}
\label{noise}
\end{figure}

Although our crossbar-aware pruning framework is effective on all testing benchmarks, some limitations still deserve further investigation. (1) The accuracy drop resulted from the pruning of deep layers cannot be significantly improved by the proposed reorder strategy. It is preferred to design a more efficient reorder method for optimizing the tolerance for pruning. (2) We observe that there are many additional crossbars not for compute but for data buffer or activation reduce, which cannot be reduced in current framework. In the future, joint optimization of the mapping and pruning frameworks is required for a better trade-off between the resource overhead and model accuracy. (3) Although our method is applicable to any crossbar architecture, some real device defects should be sufficiently considered if we use memristor-based devices (e.g. RRAM or PCRAM). For example, each device usually has limited resistance states which causes quantization error and has state variation which causes noise error. Figure 12 shows the additional analysis about influence of quantization and variation based on a pruned model. The model we choose is the VGG8 on CIFAR-10 and the sparsity we set here is 50\%. The weight with noise is governed by $we^{\theta}$, where $\theta$ is a random variable obeying a normal distribution whose mean is zero and standard deviation is the `device variation' in the figure. Obviously, the accuracy degrades along with larger variation and fewer quantized levels. In this case, $>$0.1 variation or $<$5-bit quantization will cause obvious accuracy loss. The accuracy recovery can be obtained by fabrication improvement, as well as the design of programming strategy and model retraining \cite{mohanty2017random}.

\bibliographystyle{IEEETran}
\bibliography{./ref/ref}

\begin{IEEEbiography}[{\includegraphics[width=1in,height=1.25in,clip,keepaspectratio]{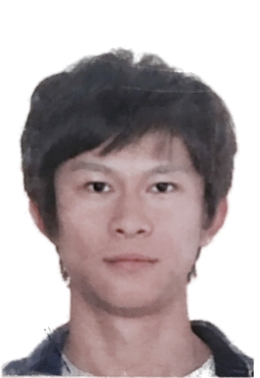}}] {Ling Liang} received the B.E. degree from Beijing University of Posts and Telecommunications, Beijing, China, in 2015, and M.S. degree from University of Southern California, CA, USA, in 2017. He is currently pursuing the Ph.D. degree at Department of Electrical and Computer Engineering,  University of California, Santa Barbara, CA, USA. His current research interests include machine learning security and computer architecture.
\end{IEEEbiography}
\vspace{-30pt}

\begin{IEEEbiography}[{\includegraphics[width=1in, height=1.25in, clip, keepaspectratio]{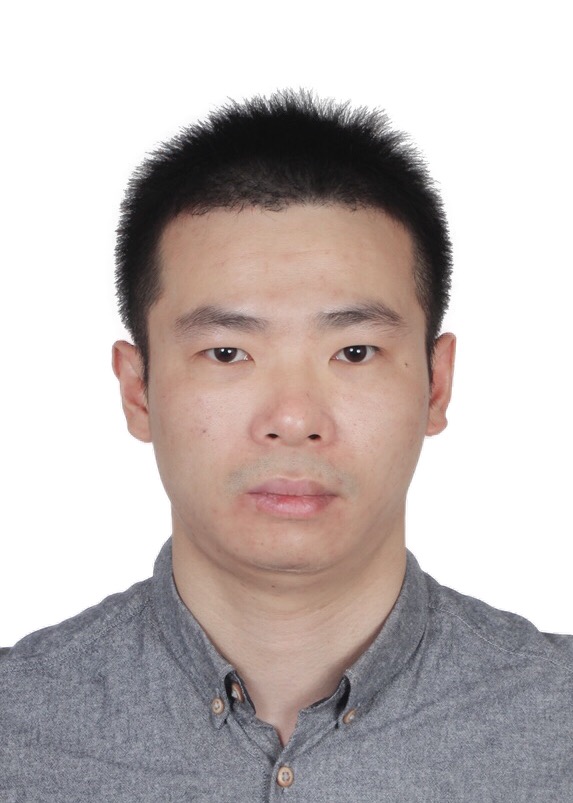}}] {Lei Deng} received his B.E. degree and Ph.D. degree from University of Science and Technology of China, Hefei, China, and Tsinghua University, Beijing, China, in 2012 and 2017, respectively. He is currently a Postdoc at Department of Electrical and Computer Engineering, University of California, Santa Barbara, CA, USA. His research interests include computer architecture, machine learning, computational neuroscience, tensor computing, and complex systems.
\end{IEEEbiography}
\vspace{-30pt}

\begin{IEEEbiography}[{\includegraphics[width=1in, height=1.25in, clip, keepaspectratio]{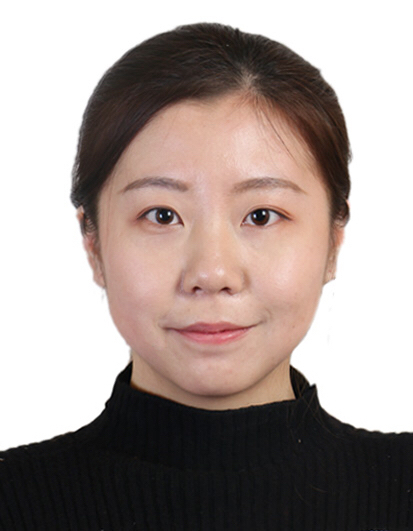}}] {Yueling Zeng} received her B.S. degree from University of California, Santa Barbara, CA, USA, in 2018. She will continue to pursue the Ph.D. degree at Department of Electrical and Computer Engineering, University of California, Santa Barbara, CA, USA. Her current research interests include machine learning and design automation.
\end{IEEEbiography}
\vspace{-30pt}

\begin{IEEEbiography}[{\includegraphics[width=1in,height=1.25in,clip,keepaspectratio]{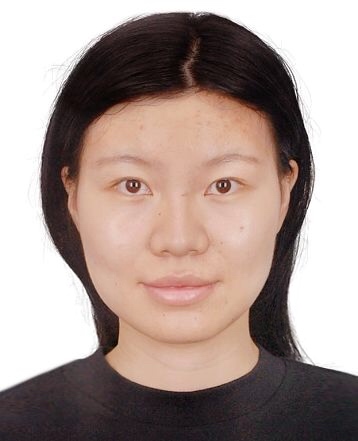}}] {Xing Hu} received the B.S. degree from Huazhong University of Science and Technology, Wuhan, China, and Ph.D. degree from University of Chinese Academy of Sciences, Beijing, China, in 2009 and 2014, respectively. She is currently a Postdoc at Department of Electrical and Computer Engineering, University of California, Santa Barbara, CA, USA. Her current research interests include emerging memory system and domain-specific hardware computing.
\end{IEEEbiography}
\vspace{-30pt}

\begin{IEEEbiography}[{\includegraphics[width=1in,height=1.25in,clip,keepaspectratio]{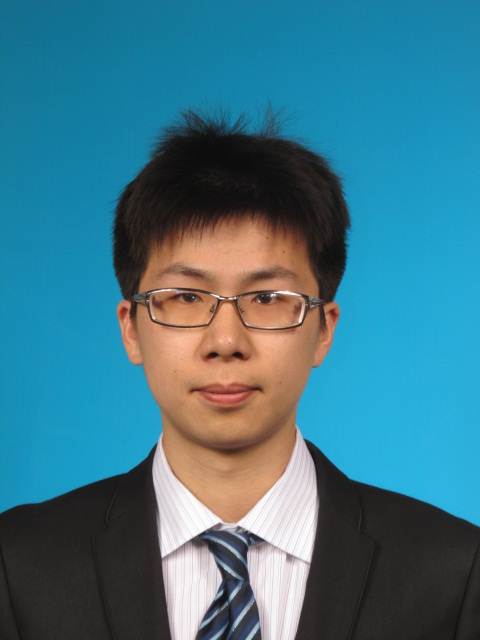}}] {Yu Ji} received his B.S. degree in physics from Tsinghua University, Beijing, China, in 2011. He is currently pursuing the Ph.D. degree at Department of Computer Science and Technology, Tsinghua University, Beijing, China. His research interests include neural network accelerator and compiler, as well as machine learning for system optimization.
\end{IEEEbiography}
\vspace{-30pt}

\begin{IEEEbiography}[{\includegraphics[width=1in, height=1.25in, clip, keepaspectratio]{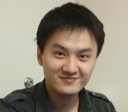}}] {Xin Ma} received his B.S. and Ph.D degree in Physics from University of Science and Technology of China, Hefei, China, in 2009 and The College of William and Mary, VA, USA, in 2014 respectively. He is currently a Postdoc at Department of Electrical and Computer Engineering, University of California, Santa Barbara, CA,USA. His research interests include designing in-memory computing with emerging non-volatile memory technologies.
\end{IEEEbiography}
\vspace{-310pt}

\begin{IEEEbiography}[{\includegraphics[width=1in,height=1.25in,clip,keepaspectratio]{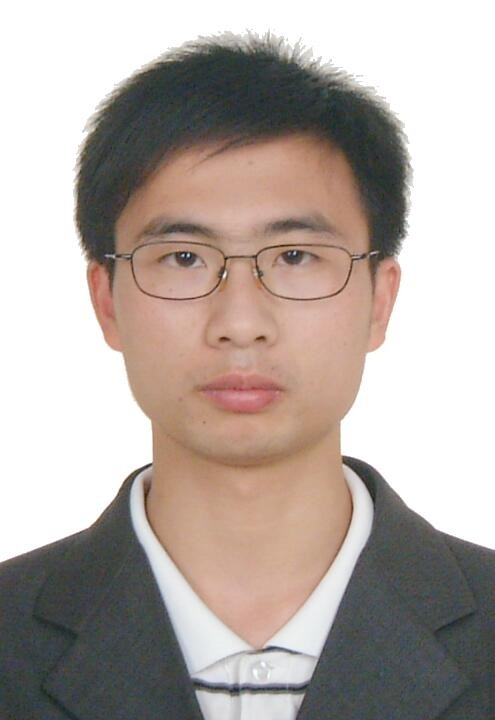}}]{Guoqi Li}  received the B.E. degree from the Xi’an University of Technology, Xi’an, China, in 2004, the M.E. degree from Xi’an Jiaotong University, Xi’an, China, in 2007, and the Ph.D. degree
from Nanyang Technological University, Singapore, in 2011. He was a Scientist with Data Storage Institute and the Institute of High Performance Computing,
Agency for Science, Technology and Research (ASTAR), Singapore, from 2011 to 2014. He is currently an Associate Professor with the Department of Precision Instrument, Tsinghua University, Beijing, China. He has published over 70 journal and conference papers. His current research interests include brain inspired computing, complex systems, neuromorphic computing, machine learning, and system identification.
\end{IEEEbiography}
\vspace{-290pt}

\begin{IEEEbiography}[{\includegraphics[width=1in,height=1.25in,clip,keepaspectratio]{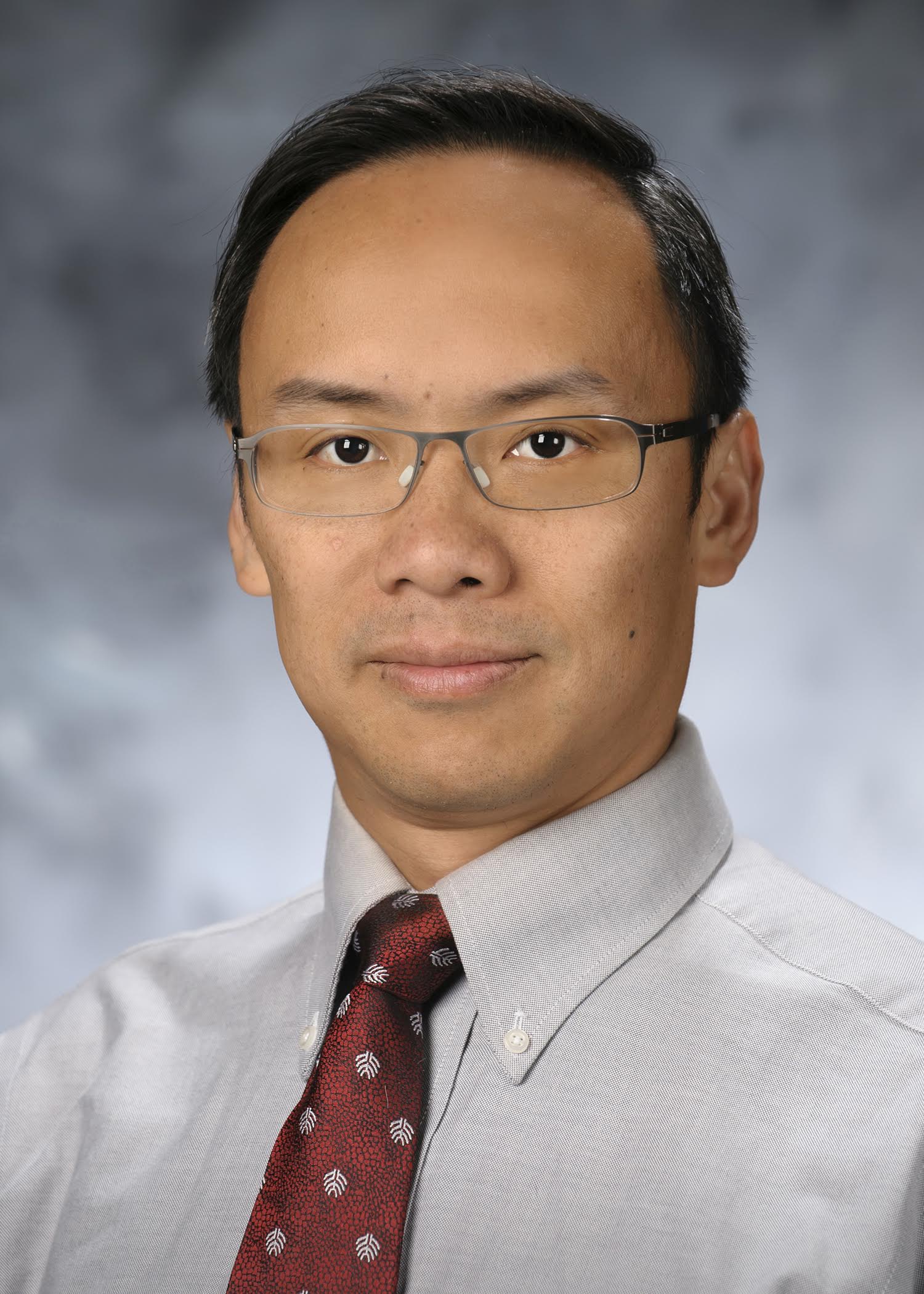}}]{Yuan Xie}  received the B.S. degree from Electrical Engineering Department, Tsinghua University, Beijing, China, in 1997, and the M.S. and Ph.D. degrees from Electrical Engineering Department, Princeton University, Princeton, NJ, USA, in 1999 and 2002, respectively. He was with IBM, Armonk, NY, USA, from 2002 to 2003, and AMD Research China Lab, Beijing, China, from 2012 to 2013. He has been a Professor with Pennsylvania State University, State College, PA, USA, since 2003. He is currently a Professor with the Electrical and Computer Engineering Department, University of California at Santa Barbara, Santa Barbara, CA, USA. He has been inducted to ISCA/MICRO/HPCA Hall of Fame. He has been an IEEE Fellow since 2015. His current research interests include computer architecture, Electronic Design Automation, and VLSI design.
\end{IEEEbiography}

\end{document}